%% file: main.tex
\PassOptionsToPackage{table}{xcolor}
\documentclass[10pt,twocolumn,letterpaper]{article}

\usepackage[pagenumbers]{iccv} 

\definecolor{iccvblue}{rgb}{0.21,0.49,0.74}
\usepackage[pagebackref,breaklinks,colorlinks,allcolors=iccvblue]{hyperref}
\definecolor{cvprblue}{rgb}{0.21,0.49,0.74}
\usepackage{makecell}
\usepackage{wrapfig}
\usepackage[table]{xcolor}
\usepackage{multirow}
\usepackage{ulem}
\def\paperID{4480} 
\def\confName{ICCV}
\def\confYear{2025}

\newcommand*{\Cline}[1]{%
\noalign{\global\setlength{\arrayrulewidth}{1pt}}%
\cline{#1}%
\noalign{\global\setlength{\arrayrulewidth}{0.4pt}}%
}

\title{IQPFR: An Image Quality Prior for Blind Face Restoration and Beyond.}
\author{\textbf{Peng Hu}$^{1,*}$\,,
\textbf{Chunming He}$^{2,}$\thanks{Equal Contribution, $\dagger$ Corresponding Author}\,,
\textbf{Lei Xu}$^3$\,, 
	\textbf{Jingduo Tian}$^4$\,,\\
        \textbf{Sina Farsiu}$^2$\,,
	\textbf{Yulun Zhang}$^5$\,,
         \textbf{Pei Liu}$^{4,\dagger}$\,,
	and \textbf{Xiu Li}$^{1,\dagger}$\\
	$^1$Shenzhen International Graduate School, Tsinghua University, 
	$^2$Duke University,\\
	$^3$Shanghai Artificial Intelligence Laboratory, 
	$^4$Media Technology Lab, Huawei, \\
	$^5$Shanghai Jiao Tong University
 }

\begin{document}
\maketitle
\input{hp_sec/0_abstract}    
\input{hp_sec/1_intro}

\input{hp_sec/2_related_work}

\input{hp_sec/3_methods}

\input{hp_sec/4_experiments}

{
    \small
    \bibliographystyle{ieeenat_fullname}
    \bibliography{main}
}
\input{hp_sec/X_suppl}

\end{document}

%% file: hp_sec/0_abstract.tex
\begin{abstract}

Blind Face Restoration (BFR) addresses the challenge of reconstructing degraded low-quality (LQ) facial images into high-quality (HQ) outputs. Conventional approaches predominantly rely on learning feature representations from ground-truth (GT) data; however, inherent imperfections in GT datasets constrain restoration performance to the mean quality level of the training data, rather than attaining maximally attainable visual quality. 
To overcome this limitation, we propose a novel framework that incorporates an Image Quality Prior (IQP) derived from No-Reference Image Quality Assessment (NR-IQA) models to guide the restoration process toward optimal HQ reconstructions. 
Our methodology synergizes this IQP with a learned codebook prior through two critical innovations: (1) During codebook learning, we devise a dual-branch codebook architecture that disentangles feature extraction into universal structural components and HQ-specific attributes, ensuring comprehensive representation of both common and high-quality facial characteristics. 
(2) In the codebook lookup stage, we implement a quality-conditioned Transformer-based framework. NR-IQA-derived quality scores act as dynamic conditioning signals to steer restoration toward the highest feasible quality standard. This score-conditioned paradigm enables plug-and-play enhancement of existing BFR architectures without modifying the original structure. 
We also formulate a discrete representation-based quality optimization strategy that circumvents over-optimization artifacts prevalent in continuous latent space approaches. 
Extensive experiments demonstrate that our method outperforms state-of-the-art techniques across multiple benchmarks. Besides, our quality-conditioned framework demonstrates consistent performance improvements when integrated with prior BFR models. The code will be released.
\end{abstract}

%% file: hp_sec/1_intro.tex
\section{Introduction}

\label{sec:intro}

Blind face restoration (BFR) aims to enhance degraded face images while preserving subject identity. This task poses significant challenges due to unknown degradations in real-world low-quality (LQ) face images. Additionally, the presence of multiple plausible high-quality (HQ) outputs for a single LQ input renders BFR an ill-posed problem.

To address the ill-posed nature of BFR, prior works \cite{chen2018fsrnet,kim2019progressive,he2024diffusion,ma2020deep,li2020blind,li2020enhanced,wang2021gfpgan,wang2023dr2,IDM,codeformer} have effectively leveraged various priors to enrich HQ details and enhance model robustness against diverse degradations. Among these, methods based on codebook priors \cite{codeformer,gu2022vqfr,wang2022restoreformer,tsai2023dual} have demonstrated promising results. These approaches frame BFR as a code prediction task within a discrete representation space, typically comprising two stages: codebook prior learning and codebook lookup. In the first stage, HQ images train a context-rich codebook via a vector quantization autoencoder \cite{vqgan}. Once the HQ codebook prior is learned, various modules and strategies (e.g., Transformers) are employed to retrieve the HQ code entries for LQ inputs in the second stage. This discrete representation framework substantially reduces restoration uncertainty, thereby improving the model's robustness to various degradation types.

\begin{figure}
  \setlength{\abovecaptionskip}{-0.25cm}
    \includegraphics[width=\linewidth,scale=0.1]{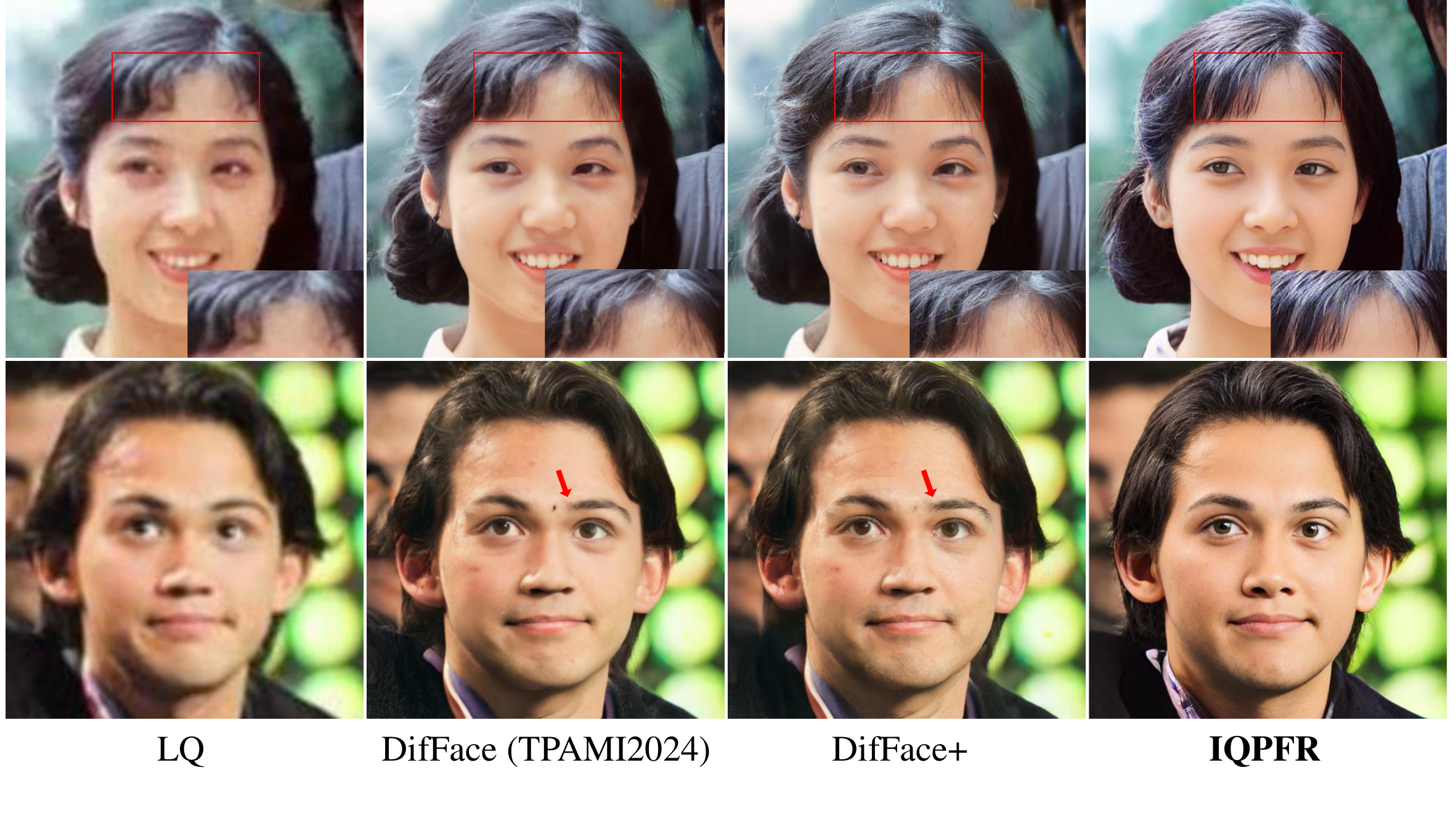}
    \caption{Results on blind face restoration. DifFace+ is the DifFace\cite{yue2022difface} with our quality prior conditioned approach, having more details. Our IQPFR has the highest perceptual quality.}
    \label{Fig.first_show}
\vspace{-0.6cm}
\end{figure}
\begin{figure}
  \setlength{\abovecaptionskip}{0cm}
    \includegraphics[width=\linewidth,scale=1]{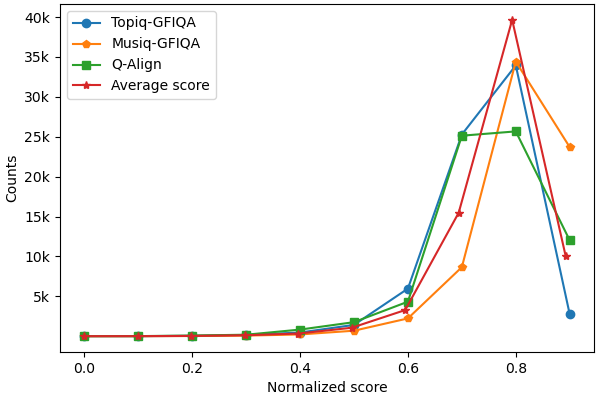}
    \caption{Score distribution of FFHQ, where higher scores correspond to better quality. Notably, most ground-truth images only exhibit an average quality score between 0.7 and 0.8, with some residual degradation patterns. This limits the network’s ability to restore low-quality inputs to truly high-quality images, achieving scores closer to 0.9. 
    }
    \label{Fig.ffhqscore}
\vspace{-0.6cm}
\end{figure}

Existing work implicitly assumes that high-quality ground truth (GT) is perfect and uses it as supervision to train restoration models, with the goal of enhancing low-quality (LQ) inputs to match the GT quality. However, these approaches often overlook the inconsistency in perceptual quality across GT data. Figure \ref{Fig.ffhqscore} shows the quality distribution of FFHQ dataset. GT data with relatively low perceptual quality can degrade the final quality of the restoration results. In other words, current methods tend to restore LQ inputs to the average quality of the GT, rather than targeting the highest achievable quality within the GT. 
Although employing NR-IQA models for data filtering (where only HQ+ images, \textit{i.e.}, those real high-quality ones whose IQA score is relatively high, are retained for training) can improve the average output quality, this approach risks compromising facial diversity since top-tier quality images typically constitute a limited subset. 
Exclusive reliance on these curated samples during training may result in facial artifacts and degraded feature representation in synthesized outputs.

In this paper, we propose leveraging an Image Quality Prior to address suboptimal restoration quality due to imperfect GT data. The image quality prior, derived from pre-trained NR-IQA models\cite{chen2023topiq,ke2021musiq,wu2023q}, captures the quality distribution of the data, enabling restoration models to distinguish images of different quality levels and to associate data with corresponding quality scores.


Specifically, we propose a novel framework, \textbf{IQPFR}, which integrates image quality prior with codebook prior to improve restoration quality. We first employ NR-IQA models to obtain image quality prior and incorporate it at different training stages. In the codebook prior learning stage, we introduce a dual-codebook architecture that comprises a common codebook and an HQ\textbf{+} codebook. The two codebooks are trained jointly but with distinct data flows: the common codebook learns features from the entire dataset, while the HQ\textbf{+} codebook learns exclusively from a subset of the highest-quality data (HQ\textbf{+} images). In the code prediction stage, we introduce a quality-prior-conditioned Transformer that uses the quality score as a condition during code prediction. This conditional Transformer can perceive the quality level of recovered images, enabling the generation of face images with the highest quality when provided with the highest score as input condition.
Importantly, the feasibility of this score-conditioned injection lies in the inconsistent quality of the training data, making it applicable to other BFR models. Additionally, we utilize the discrete representation space provided by the codebook for quality optimization. Compared to continuous representations, the discrete representation significantly mitigates the over-optimization issue \cite{gao2023scaling}. The maximization of the quality score directs IQPFR to generate the highest-quality images possible.
Figure \ref{Fig.first_show} shows the effectiveness of our IQPFR and the quality prior conditioned approach.


In summary, our main contributions are as follows:
\begin{itemize}
\item We propose the use of image quality prior to aid the restoration model in understanding HQ features, thereby enhancing its capacity to generate HQ results. To the best of our knowledge, this is the first work to bridge the gap between image quality prior and its application in improving restoration quality for blind face restoration.  

\item We introduce a series of strategies for integrating image quality prior with codebook prior, including a dual-codebook reconstruction model, a plug-and-play score-conditioned approach that uses the image quality score as a condition, and a quality optimization technique based on discrete representation. These techniques leverage the advantages of image quality prior across multiple stages to improve restoration quality. 

\item Extensive experiments demonstrate that our IQPFR significantly surpasses previous works in terms of restoration quality and our proposed score-conditioned approach functions as a plug-and-play manner to enhance the quality of existing BFR methods.
\end{itemize}

%% file: hp_sec/2_related_work.tex
\section{Related Work}

\subsection{Blind face restoration}
Various priors have been explored to support high-quality generation and robust identity preservation in blind face restoration (BFR). These priors are generally classified into three types: geometric priors \cite{chen2018fsrnet,kim2019progressive,ma2020deep,chen2021progressive,hu2020face,hu2021face}, reference priors \cite{gfrnet,li2020blind,li2020enhanced,dogan2019exemplar}, and generative priors \cite{wang2021gfpgan,he2022gcfsr,wang2023dr2,IDM,wang2022panini}.

Geometric priors typically extract structural information from low-quality face images, such as facial landmarks \cite{chen2018fsrnet,kim2019progressive,ma2020deep}, 3D shapes \cite{hu2020face,hu2021face}, and facial component heatmaps \cite{chen2021progressive}. However, geometric priors often fail to provide sufficient detail for accurate restoration, as it is difficult to reliably extract features from severely degraded images.
Reference priors leverage high-quality images with the same identity but different content from the degraded image as a guide \cite{li2020blind}. A major limitation of this approach is the challenge of obtaining high-quality reference images of the same identity as the degraded input.
With advances in generative models, several works have employed generative priors to supply detailed facial features for BFR \cite{wang2021gfpgan,wang2023dr2,IDM,lin2023diffbir,qiu2023diffbfr,li2025interlcm}. For instance, DR2 \cite{wang2023dr2} utilizes a pre-trained diffusion model alongside an enhancement module to restore degraded face images. While generative priors can introduce diverse facial details, they often struggle to maintain high fidelity without additional guidance.

Codebook priors leverage a compressed codebook to store high-quality features, which are subsequently referenced during the restoration of low-quality inputs. Current works \cite{gu2022vqfr,codeformer,wang2022restoreformer,tsai2023dual} generally involve two stages: codebook prior learning and codebook lookup. During the codebook learning stage, a high-quality codebook is constructed using a vector quantized autoencoder (e.g., VQGAN \cite{vqgan}), which automatically stores optimal high-quality features without requiring a handcrafted dictionary. In the lookup stage, each approach employs tailored strategies, such as the parallel decoder in VQFR \cite{gu2022vqfr}, multi-head cross-attention in RestoreFormer \cite{wang2022restoreformer}, and Transformers in CodeFormer \cite{codeformer} and DAEFR \cite{tsai2023dual}. Both CodeFormer and DAEFR freeze the codebook and decoder during the lookup stage, ensuring that the output space remains discrete. This approach reduces restoration uncertainty, giving codebook priors an inherent advantage in maintaining fidelity compared to models operating within a continuous space.

\begin{figure*}[!htbp]
  \centering
    \setlength{\abovecaptionskip}{-0.0cm}
  \includegraphics[scale=1.01]{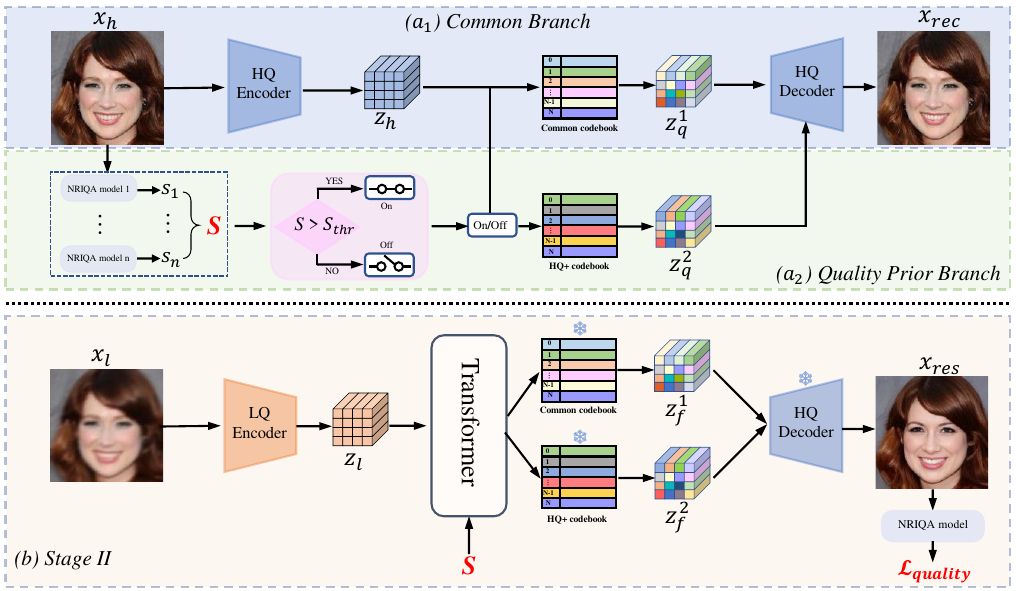}
  \caption{\textbf{Overall framework of IQPFR.} (a) In the codebook learning stage, a dual-codebook architecture is proposed. The HQ\textbf{+} codebook is learned to quantize $Z_h$ only when the quality score of $x_h$ is higher than the threshold $S_{thr}$.
  (b) In the codebook lookup stage, we input the quality score $S$ as a condition into Transformer \textit{T}, which predicts two code sequences at the same time. 
  The two codebooks are leveraged to look up the corresponding code entries. Finally, the NR-IQA model is utilized to calculate the quality loss $\mathcal{L}_{quality}$.
  }
  \vspace{-0.4cm}
  \label{fig:architecture}
\end{figure*}
\subsection{Image Quality Prior}
Non-reference image quality assessment (NR-IQA) models \cite{ke2021musiq,chen2023topiq,qin2023data,agnolucci2024arniqa,wang2023exploring,ni2024opinion,he2025reti,he2023hqg} enable estimation of quality distribution for input images without the need for any reference images. Leveraging this quality distribution as a prior, a restoration model can associate images with their respective quality levels, enhancing its capacity to accurately perceive restoration quality. For instance, EnhanceGAN \cite{enhancegan} employs a binary classification discriminator (high-quality or low-quality) to guide its generator for image enhancement, while DOODL \cite{wallace2023end} utilizes an aesthetic scoring model as classifier guidance to optimize latent diffusion, thereby improving the aesthetic quality of images.
However, these methods only minimally explore the potential of quality priors, generally relegating them to a supporting role. By contrast, we have developed three distinct applications of image quality priors specifically for blind face restoration (BFR), each tailored to the unique characteristics of quality priors and the demands of the BFR task. These designs maximize the utility of image quality priors across various stages, significantly enhancing restoration quality.

%% file: hp_sec/3_methods.tex
\setlength{\abovedisplayskip}{3pt}
\setlength{\belowdisplayskip}{3pt}

\section{Methodology}
To achieve blind face restoration (BFR) with both high quality and high fidelity, we introduce IQPFR, a framework that integrates image quality priors into both the codebook learning and codebook lookup stages through distinct strategies. The overall framework of IQPFR is depicted in~\Cref{fig:architecture}.
In the codebook learning stage, we employ a dual-codebook architecture designed to retain diverse and detailed facial features (\Cref{subsec:3.1}). In the codebook lookup stage, this dual codebook is utilized alongside a quality-prior conditional Transformer network for code prediction (\Cref{3.2}). Additionally, image quality is incorporated as a training objective to further guide and enhance restoration quality (\Cref{3.3}).

\subsection{Dual Codebook Learning}
\label{subsec:3.1}
\paragraph{Preliminary.}
The discrete codebook is learned through a self-reconstruction process initially introduced in VQVAE \cite{van2017neural} and later enhanced in VQGAN \cite{vqgan}, which incorporates adversarial and perceptual loss objectives to improve perceptual quality. Specifically, VQGAN consists of three modules: an encoder (\textit{E}), a codebook (\textit{C}), and a decoder (\textit{D}). For an input image $\textit{x}_h \in \mathbb{R}^{H \times W \times 3}$, the encoder extracts a latent feature map $Z_h = \textit{E}(\textit{x}_h) \in \mathbb{R}^{h \times w \times c}$, where $c$ represents the dimensionality of both $Z_h$ and the code vectors within the codebook. Next, for each of the $h \times w$ vectors in $Z_h$, we compute the representation distance to the entries in the codebook $\textit{C}$ and map each vector to the closest codebook entry, thereby obtaining the vector-quantized representation $Z_q \in \mathbb{R}^{h \times w \times c}$: 
\begin{equation}
  Z_q^{(i,j)} = \mathop{\arg\min}_{C_k \in \textit{C}}\| Z_h^{(i,j)}-C_k\|_2,
  \label{eq:zq}
\end{equation}
where $C_k$ represents the codebook entries in $\textit{C}$.
Given $Z_q$, the decoder $\textit{D}$ reconstruct the HQ image $\textit{x}_{rec} \in \mathbb{R}^{H\times W \times 3}$. 

\paragraph{Dual codebook.}
We recognize that the quality of inputs can vary significantly, with those inputs of the highest quality containing more valuable facial details. Consequently, we propose a dual codebook architecture that includes a common codebook and a high-quality (HQ\textbf{+}) codebook.
As depicted in ~\Cref{fig:architecture}(a), the common codebook is learned from all HQ inputs, while the HQ\textbf{+} codebook is learned exclusively from a subset of images that exceed a certain quality threshold $S_{thr}$. 
Specifically, we first use the common codebook to quantize $\textit{E}(x_h)$, resulting in $Z_q^1$. Then we employ NR-IQA models to assess the quality score of all inputs. 
If the average quality score of the current HQ input, denoted as $S$, surpasses the threshold $S_{thr}$, we then engage the HQ\textbf{+} codebook to quantize $\textit{E}(x_h)$, yielding $Z_q^2$. $Z_q$ is formulated as follows:
\begin{equation}
  Z_q=
  \begin{cases}
    Z_q^1+\alpha Z_q^2  & if \text{ } S>S_{thr} ,  \\
    Z_q^1 & if \text{ } S <= S_{thr},
  \end{cases}
  \label{eq:dual_codebook}
\end{equation}
where $\alpha$ is a balance weight. In cases where the quality threshold is not met, only the common codebook is employed. 
The decoder $\textit{D}$ then receives $Z_q$ as input and outputs the reconstructed face image $x_{rec}$. 

The common codebook, trained on a diverse range of face images, encompasses a wide variety of facial features, while the HQ\textbf{+} codebook captures high-quality features specific to the HQ\textbf{+} images. The decoder \textit{D} learns to reconstruct HQ\textbf{+} image when receiving the fusion of common feature $Z_q^1$ and HQ\textbf{+} feature $Z_q^2$. During the inference phase, we match the HQ\textbf{+} feature for $x_h$ and integrate it with the common feature, allowing \textit{D} to generate a HQ\textbf{+} image $x_{res}$, even if $x_h$ is a common image.

\paragraph{Training objective.} Following the practice of existing methods~\cite{codeformer,tsai2023dual},
our training objectives employ reconstruction L1 loss $\mathcal{L}_1 $, perceptual loss $\mathcal{L}_{per}$, adversarial loss $\mathcal{L}_{adv}$ and code feature loss $\mathcal{L}_{feat}$ for optimization:
\begin{equation}
  \centering
  \begin{aligned}
    \mathcal{L}_1&={\|\textit{x}_h - \textit{x}_{rec}\|}_1; 
        \label{eq:loss_l1}
  \end{aligned}
\end{equation}
\begin{equation}
  \centering
  \begin{aligned}
    \mathcal{L}_{per}={\|\Phi(\textit{x}_h)-\Phi(\textit{x}_{rec}) \|}_2^2; 
            \label{eq:loss_lper}
  \end{aligned}
\end{equation}
\begin{equation}
  \centering
  \begin{aligned}
    \mathcal{L}_{adv}=log\mathcal{D}(x_{h}) +log(1-\mathcal{D}(x_{rec})); 
            \label{eq:loss_ladv}
  \end{aligned}
\end{equation}
\begin{equation}
  \centering
  \begin{aligned}
    \mathcal{L}&_{feat}=\|sg(Z_h)-Z_q\|_2^2+ \beta \|Z_h-sg(Z_q)\|_2^2,
    \label{eq:loss_lfeat}
  \end{aligned}
\end{equation}
where $\Phi$ is the feature extractor of VGG19\cite{simonyan2014very}, $\mathcal{D}$ is the discriminator \cite{isola2017image} and sg($\cdot$) denotes the stop-gradient operator.
Considering the first item of $\mathcal{L}_{feat}$,
if $S > S_{thr}$, $Z_q = Z_q^1+Z_q^2$ is guided to approximate the HQ\textbf{+} image features, otherwise $Z_q=Z_q^1$ is guided to approximate the common image features, so $Z_q^2$ can be regarded as the residual between the HQ\textbf{+} images and the common images.

\subsection{Quality Prior Condition}
\label{3.2}
In the codebook lookup stage, we introduce a quality prior as an additional condition for code prediction using a conditional Transformer. This Transformer model predicts two code sequences conditioned on the image quality score $S$, which represents the quality of the corresponding GT image as determined by NR-IQA models in the codebook learning stage. The conditional Transformer is trained to capture the relationship between GT images and their associated quality scores. Specifically, during training, when the highest quality score is used as the condition, the ground truth is set to the highest quality image. Thus, after training, inputting the highest score as a condition prompts the model to produce restoration results with the highest achievable quality.

First, we extract the LQ feature via $Z_l = \textit{E}(x_l)$, 
then we embed the quality score $s$ into an embedding vector $\textbf{s} \in \mathbb{R}^{h*w*c} $ and reshape it to $\textbf{s} \in \mathbb{R}^{h \times w \times c} $, which matches the dimensionality of $Z_l$. 
This embedding vector $\textbf{s}$ is directly added  to $Z_l$ as follows:
\begin{equation}
  \centering
  \begin{aligned}
      \widehat{Z}_l=Z_l+\textbf{s}.
    \label{eq:z1_z2}
  \end{aligned}
\end{equation}
Upon receiving $\widehat{Z}_l$ as input, the Transformer $\textit{T}$ predicts two code sequences, $\textbf{c}_1$ and $\textbf{c}_2$.
Then $\textbf{c}_1$ retrieves code entries from the common codebook to construct the quantized feature $Z_f^1$, while $\textbf{c}_2$ retrieves code entries from the HQ\textbf{+} codebook to form  $Z_f^2$.
Then we fuse $Z_f^1$ and $Z_f^2$ to get $Z_f$ via 
\begin{equation}
  \centering
  \begin{aligned}
      Z_f=Z_f^1+\alpha Z_f^2,
    \label{eq:zf}
  \end{aligned}
\end{equation}
where $\alpha$ is a balance weight, same as ~\Cref{eq:dual_codebook}. And $Z_f$ is fed to the decoder to generate restoration images $x_{res}$. Since \textit{D} has learned to reconstruct HQ\textbf{+}
image from the fusion of $Z_q^1$ and $Z_q^2$, the generated image in Stage \uppercase\expandafter{\romannumeral2} has a similar quality to HQ\textbf{+} image when \textit{D} takes $Z_f$ as inputs.

With the quality prior as a condition, we can control the quality of recovered images during inference. Typically, the objective is to achieve the highest possible quality, so we input the maximum quality score as the condition. The Transformer then generates code sequences that, upon decoding, yield an image of the highest achievable quality.

\noindent\textbf{Quality Prior Ensembles.}
Although existing NR-IQA methods are highly powerful and continue to advance, they may still harbor certain biases. Directly using NR-IQA to guide restoration models could risk transferring these biases into the restoration process. For example, if an IQA model has a particular preference for blue eyes, a restoration model guided by it might be inclined to produce high-quality restored face images with blue eyes.

To mitigate the impact of biases introduced by a single IQA, we propose to incorporate multiple IQA models as the final IQA prior \cite{coste2023reward}. Specifically, we calculate the mean score of different IQA models:
\begin{equation}
  \centering
  \begin{aligned}
    S=\frac{1}{n}\sum_{i=1}^{n}s_{i},
    \label{eq:score ensemble}
  \end{aligned}
\end{equation}
where $s_{i}$ is the normalized score of the $i$-th IQA model. This ensemble method is used in both the codebook learning stage and the code prediction stage.



\subsection{Quality Optimization}
\label{3.3}
Using IQA measures as objectives in image processing systems represents a promising but under-explored area, especially in the context of NR-IQA methods. Prior studies have investigated FR-IQA measures as objectives \cite{channappayya2008ssim,snell2017learning,johnson2016perceptual,ding2021comparison}. In theory, employing NR-IQA objectives could enhance the perceptual quality of image processing outputs. However, unlike FR-IQA measures, the optimization direction of NR-IQA objectives is unstable due to the absence of a reference.

This instability introduces a significant limitation when using NR-IQA measures as objectives: restoration models can produce images with low perceptual quality from a human perspective, yet deceive NR-IQA models into assigning high IQA scores. Such images are known as \emph{adversarial examples} \cite{ding2021comparison}. In optimization, model parameters are tuned to maximize the IQA score of generated images, which can inadvertently lead the restoration model to generate adversarial examples against NR-IQA models. Similar issues have been observed in reward optimization of Large Language Models~\cite{gao2023scaling}, where this effect is termed over-optimization. Over-optimization arises because IQA models serve only as proxies for human preferences, and thus may not perfectly align with actual human perceptual standards. Consequently, optimization may not reliably enhance perceptual quality from a human viewpoint.

The presence of adversarial examples renders NR-IQA objectives unreliable. A fundamental solution involves reducing the presence of adversarial examples within the output space. If adversarial samples are eliminated from the output space, NR-IQA scores can be maximized with confidence. There are two strategies to achieve this: (1) Improving the consistency of NR-IQA scores with human preferences. 
(2) Limiting the size and distribution of output space.
We demonstrate that a discrete codebook prior naturally supports the second strategy. Unlike continuous priors, the discrete codebook prior restricts the output space to a finite set, significantly reducing the number of adversarial examples. Additionally, training the codebook prior on HQ face images ensures that most samples within this finite space possess high-quality semantic facial information, thereby limiting the proportion of adversarial examples.

Thus, NR-IQA measures are particularly well-suited as objectives for discrete codebook-based network structures. Based on this, we propose to maximize NR-IQA scores directly to fine-tune our restoration model parameters, which further enhances the restoration quality.
 
\noindent\textbf{Final training objective.} 
In the second stage, three losses are utilized to train the encoder $\textit{E}$ and the Transformer $\textit{T}$: an L2 loss $\mathcal{L}_{feat} $, a cross-entropy loss $\mathcal{L}_{index}$ and a quality loss $\mathcal{L}_{quality}$.
\begin{equation}
  \centering
  \begin{aligned}
      \mathcal{L}_{feat}=\|Z_l-sg(Z_q)\|_2^2, 
    \label{eq:loss_lfeat2}
  \end{aligned}
\end{equation}
\begin{equation}
  \centering
  \begin{aligned}
      \mathcal{L}_{index}= \sum_{i=0}^{h*w-1}-\textbf{c}_1^i log(\widehat{\textbf{c}}_1^i) +\sum_{i=0}^{h*w-1}-\textbf{c}_2^i log(\widehat{\textbf{c}}_2^i),
    \label{eq:loss_lindex}
  \end{aligned}
\end{equation}
\begin{equation}
  \centering
  \begin{aligned}
\mathcal{L}_{quality}=-IQA(x_{res}).
    \label{eq:loss_quality}
  \end{aligned}
\end{equation}

 \begin{table*}[htbp]
\setlength{\abovecaptionskip}{0cm}
  \centering
  \resizebox{2.05\columnwidth}{!}{
  \renewcommand\arraystretch{1.05}
  \begin{tabular}{c|c|cccccccc}
    \Xhline{1.2pt}
    \rowcolor{gray!30}
    Datasets&Methods & TOPIQ-G$\uparrow$& Musiq-G$\uparrow$ & Musiq-K$\uparrow$  & Musiq-A$\uparrow$ &Arniqa $\uparrow $&MDFS $\downarrow$ &Q-Align$\uparrow$ &CLIP-IQA$\uparrow$\\
    \Xhline{0.7pt}
    \multirow{8}{*}{LFW-Test}
          &RestoreFormer\cite{wang2022restoreformer}&0.793&  0.807 &73.70 &4.65 &0.711&18.64&4.22&0.741\\
          &DR2\cite{wang2023dr2}&0.720 &0.753 &67.13 &4.67 &0.700&18.43&4.23&0.658\\
          &CodeFormer\cite{codeformer} &0.809&0.832& 75.47 &4.76 &0.726 &18.46&4.31&0.697\\
          &DifFace\cite{yue2022difface} &0.718&0.748&69.90&4.48&0.686&18.41&3.86&0.610\\
          &DifFace+\cite{yue2022difface} &0.792 (+0.074)&0.825 (+0.077)&74.56 (+4.66)&4.79 (+0.31)&0.725 (+0.039)&18.08 (+0.33)&\textcolor{red}{\textbf{4.73 (+0.87)}}&0.664 (+0.054)\\
          &DAEFR\cite{tsai2023dual}&0.814&0.827 & 75.84&4.81&0.742&\textcolor{blue}{\textbf{17.91}}&4.33&0.696\\
          &DAEFR+\cite{tsai2023dual} &0.821 (+0.007)&\textcolor{blue}{\textbf{0.841 (+0.014)}}&\textcolor{blue}{\textbf{76.20 (+0.36)}}&\textcolor{blue}{\textbf{4.89 (+0.08)}}&0.743 (+0.001)&\textcolor{red}{\textbf{17.82 (+0.09)}}&4.40 (+0.07)&0.710 (+0.014)\\          &WaveFace\cite{miao2024waveface}&0.786&0.799&73.55&4.69&0.689&19.18&4.43&\textcolor{red}{\textbf{0.788}}\\
          &Interlcm\cite{li2025interlcm}&\textcolor{blue}{\textbf{0.831}}&0.834&75.87&4.84&\textcolor{blue}{\textbf{0.753}}&17.98&4.55&0.721\\
          &\textbf{Ours} &\textcolor{red}{\textbf{0.859}}&\textcolor{red}{\textbf{0.876}}&\textcolor{red}{\textbf{76.87}}&\textcolor{red}{\textbf{5.06}}&\textcolor{red}{\textbf{0.758}}&17.95&\textcolor{blue}{\textbf{4.64}}&\textcolor{blue}{\textbf{0.761}}\\
  \hline
  \multirow{8}{*}{WebPhoto-Test}
          &RestoreFormer\cite{wang2022restoreformer}&0.706& 0.721 &69.83 &4.55 &0.677&19.21&3.52&0.711 \\
          &DR2\cite{wang2023dr2}&0.621 & 0.630&61.28 &\textcolor{blue}{\textbf{4.88}}&0.601&20.51&2.87&0.447\\

          &CodeFormer\cite{codeformer}&0.756 &0.782 &73.56 &4.69&0.687&19.10&3.84&0.691 \\
          &DifFace\cite{yue2022difface} &0.638&0.670&65.77&4.44&0.642&19.36&3.31&0.586\\
          &DifFace+\cite{yue2022difface} &0.734 (+0.096)&0.767 (+0.097)&72.74 (+6.97)&4.73 (+0.29)&0.677 (+0.035)&18.80 (+0.56)&3.75 (+0.44)&0.646 (+0.060)\\
          &DAEFR\cite{tsai2023dual}&0.746 &0.753 &72.70 &4.58 &0.701&18.63&3.82&0.669\\
          &DAEFR+\cite{tsai2023dual} &0.777 (+0.031)&0.788 (+0.035)&74.44 (+1.74)&4.84 (+0.26)&0.708 (+0.007)&\textcolor{blue}{\textbf{18.43 (+0.20)}}&3.96 (+0.14)&0.689 (+0.020)\\
          &WaveFace\cite{miao2024waveface}&0.694&0.704&70.46&4.48&0.686&19.66&3.76&\textcolor{red}{\textbf{0.778}}\\&Interlcm\cite{li2025interlcm}&\textcolor{blue}{\textbf{0.794}}&\textcolor{blue}{\textbf{0.807}}&\textcolor{blue}{\textbf{75.82}}&4.81&\textcolor{blue}{\textbf{0.731}}&18.45&\textcolor{blue}{\textbf{3.99}}&0.752\\
        
        &\textbf{Ours} &\textcolor{red}{\textbf{0.820}}&\textcolor{red}{\textbf{0.848}}&\textcolor{red}{\textbf{76.80}}&\textcolor{red}{\textbf{4.95}}&\textcolor{red}{\textbf{0.744}}&\textcolor{red}{\textbf{18.32}}&\textcolor{red}{\textbf{4.21}}&\textcolor{blue}{\textbf{0.760}}\\

  \hline
  \multirow{8}{*}{WIDER-Test}
          &RestoreFormer\cite{wang2022restoreformer}&0.714 & 0.733& 67.83&4.49 &0.691 &19.04&3.55&0.727\\
          &DR2\cite{wang2023dr2}& 0.731&0.758 &67.76 &4.73 &0.601 &20.06&2.87&0.523\\
          &CodeFormer\cite{codeformer} &0.772&0.810&72.97 &4.57&0.705&18.67&4.05&0.697 \\
          &DifFace\cite{yue2022difface} &0.684&0.715&65.04&4.36&0.644&18.88&3.60&0.597\\
          &DifFace+\cite{yue2022difface} &0.774 (+0.090)&0.808 (+0.093)&71.74 (+6.70)&4.66 (+0.30)&0.686 (+0.042)&18.45 (+0.43)&4.17 (+0.57)&0.653 (+0.056)\\
          &DAEFR\cite{tsai2023dual}&0.787 &0.807&74.15  &4.60 &0.721&18.18&4.17&0.697\\
          &DAEFR+\cite{tsai2023dual} &\textcolor{blue}{\textbf{0.818 (+0.031)}}&\textcolor{blue}{\textbf{0.835 (+0.028)}}&75.11 (+0.96)&4.74 (+0.14)&0.727 (+0.006)&\textcolor{red}{\textbf{18.06 (+0.12)}}&\textcolor{blue}{\textbf{4.31 (+0.14)}}&0.719 (+0.022)\\
          &WaveFace\cite{miao2024waveface}&0.751&0.778&72.90&4.70&0.686&19.81&4.12&\textcolor{red}{\textbf{0.781}}\\&Interlcm\cite{li2025interlcm}&0.798&0.820&\textcolor{blue}{\textbf{75.34}}&\textcolor{blue}{\textbf{4.80}}&\textcolor{blue}{\textbf{0.743}}&18.26&4.24&0.754\\
          
          &\textbf{Ours} &\textcolor{red}{\textbf{0.839}}&\textcolor{red}{\textbf{0.870}}&\textcolor{red}{\textbf{76.39}}&\textcolor{red}{\textbf{4.90}}&\textcolor{red}{\textbf{0.745}}&\textcolor{blue}{\textbf{18.12}}&\textcolor{red}{\textbf{4.47}}&\textcolor{blue}{\textbf{0.772}}\\
  \Xhline{1.2pt}
  \end{tabular}
  }
  \caption{Quantitative comparison on the real world datasets. 
    \label{tab:real_world_test}
  \textcolor{red}{Red} and \textcolor{blue}{blue} indicate the best and second best performance, respectively. For DifFace+ and DAEFR+, the numbers in `()' indicate the improvement for their baseline.
  }
\end{table*}

\begin{figure*}[htbp]
  \setlength{\abovecaptionskip}{-0.3cm}
  \centering
    \includegraphics[scale=0.24]{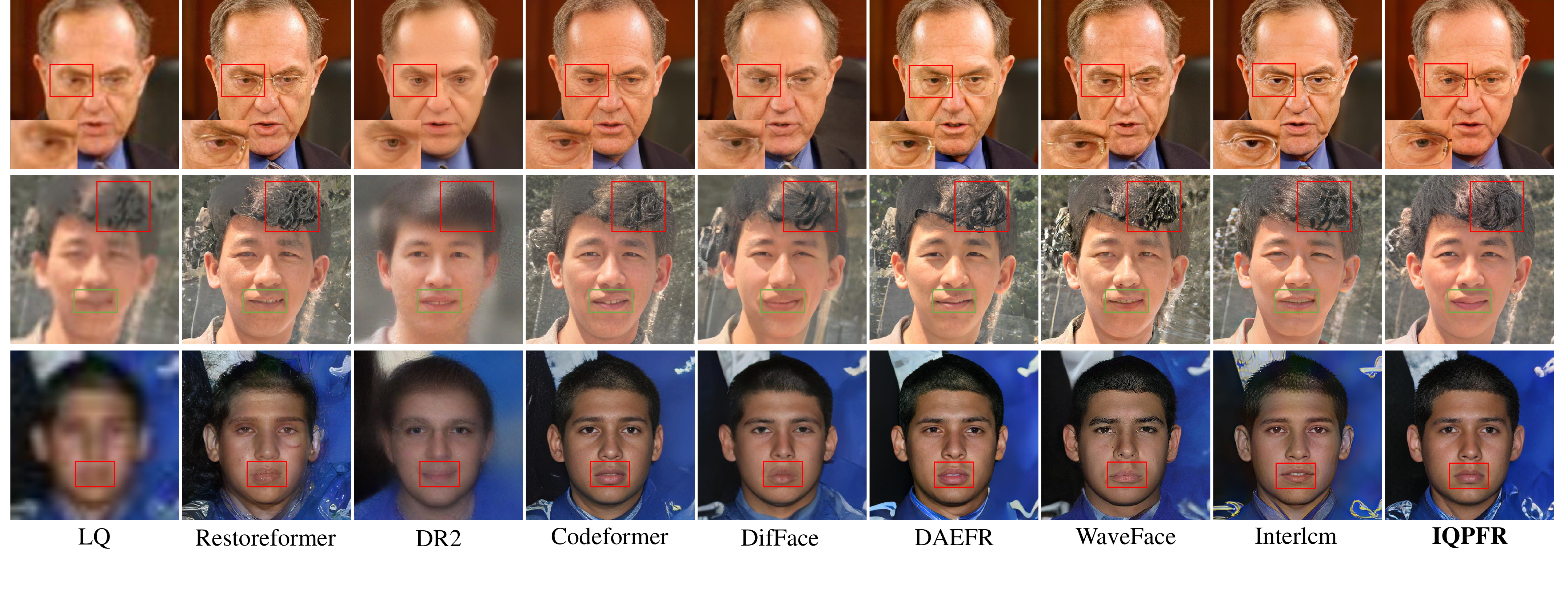}
    \caption{Qualitative comparison on real-world datasets with different degrees of degradation: LFW (first row), WebPhoto-Test (second row), WIDER-Test (third row). Our results have better facial feature and more textures.  Zoom in to see the details.}
    \label{Fig.real_world}
\vspace{-0.4cm}
\end{figure*}

$\mathcal{L}_{index}$ takes the HQ code indices $\textbf{c}_1$ and $\textbf{c}_2$ from two codebooks respectively as ground-truth, which enables \textit{T} to predict the correct code indices and thus generate the corresponding HQ images after decoding.
For $\mathcal{L}_{feat}$, the ground-truth $Z_q$ is obtained via:
\begin{equation}
  \centering
  \begin{aligned}
      Z_q=Z_q^1+\alpha Z_q^2,
    \label{eq:stage2_gt}
  \end{aligned}
\end{equation}
where $Z_q^1$ and $Z_q^2$ are retrieved from the two codebooks using $\textbf{c}_1$ and $\textbf{c}_2$ respectively.
$Z_q$ is calculated in the same way as~\Cref{eq:dual_codebook} when $S>S_{thr}$ (i.e., HQ\textbf{+} images), so $\mathcal{L}_{feat}$ is actually encouraging the LQ feature to resemble the HQ\textbf{+} feature.
Leveraging the codebook prior’s inherent restriction against adversarial examples, we propose directly using the NR-IQA score as a training objective, defined as $\mathcal{L}_{quality}=-IQA(x_{res})$. 

The overall training objective is thus formulated as:
\begin{equation}
  \centering
  \begin{aligned}
      \mathcal{L}_{f}=\mathcal{L}_{feat}+\lambda_1 \mathcal{L}_{index}+\lambda_2 \mathcal{L}_{quality}, 
    \label{eq:loss_stage2}
  \end{aligned}
\end{equation}
where $\lambda_1$ and $\lambda_2$ are set to 0.5 and 0.1, respectively.

%% file: hp_sec/4_experiments.tex
\section{Experiments}
\noindent\textbf{Implementation Details.}
\label{sec:4.1}
For our quality priors, we employ an ensemble of three advanced NR-IQA models, including both traditional and MLLM-based models, to obtain the final quality score. 
For the traditional quality prior, we use Musiq \cite{ke2021musiq} and TOPIQ \cite{chen2023topiq} trained on the GFIQA \cite{su2023going} database,
and refer to them as Musiq-GFIQA and TOPIQ-GFIQA. 
For MLLM-based models, we use Q-Align \cite{wu2023q} to assess image quality.

\begin{figure*}[!t]
  \setlength{\abovecaptionskip}{-0.1cm}
  \centering
    \includegraphics[scale=0.24]{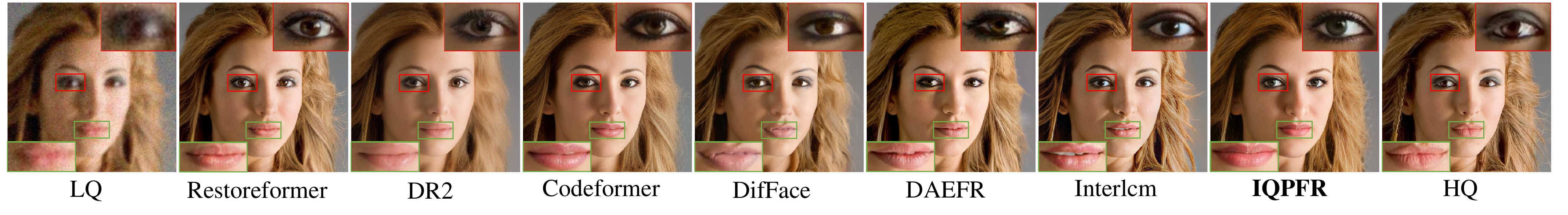}
    \caption{Qualitative comparison on Celeba-Test. Our IQPFR demonstrates superior perceptual quality with aesthetic skin tone and enhanced detail preservation, notably in fine facial features like eyelashes.}
    \label{Fig.celeba}
\vspace{-0.4cm}
\end{figure*}
\noindent\textbf{Datasets and metrics.}
We train our model on the FFHQ dataset \cite{karras2019style} with 70,000 high-quality face images. 
During the second-stage training, we follow the common practice \cite{codeformer,tsai2023dual} and manually degrade each high-quality image 
to generate HQ-LQ pairs (See appendix for detailed degradation operations).  Following \cite{wang2021gfpgan,codeformer,wang2023dr2}, 
we use a synthetic dataset, CelebA-Test, alongside three real-world datasets that exhibit varying degrees of degradation: LFW-Test \cite{huang2008labeled} (mild degradation), WebPhoto-Test \cite{wang2021gfpgan} (medium degradation), and WIDER-Test \cite{codeformer} (heavy degradation).
We apply many advanced NR-IQA metrics to evaluate the perceptual quality of restored images, including Musiq-Koniq \cite{ke2021musiq}, Musiq-GFIQA \cite{ke2021musiq}, and TOPIQ-GFIQA \cite{chen2023topiq},  Musiq-AVA \cite{ke2021musiq} (for aesthetic quality), Arniqa \cite{agnolucci2024arniqa}, MDFS \cite{ni2024opinion}, CLIP-IQA \cite{wang2023exploring} and Q-Align \cite{wu2023q}. 

\begin{table}[bp]

\vspace{-0.4cm}
\setlength{\abovecaptionskip}{0.2cm}

    \flushleft
    \renewcommand\arraystretch{3.5}
    \resizebox{1\columnwidth}{!}{
    \begin{tabular}{c|cccccccc}
      \Xhline{1.2pt}
      \rowcolor{gray!30}
      \Huge Methods & \Huge TOPIQ-G$\uparrow$& \Huge Musiq-G$\uparrow$&\Huge Musiq-K$\uparrow$  &\Huge Musiq-A$\uparrow$&\Huge Arniqa$\uparrow$&\Huge MDFS$\downarrow$&\Huge Q-Align$\uparrow$&\Huge CLIP-IQA$\uparrow$ \\[0.5em]
      \Cline{1-9}
          \Huge Input &{\Huge 0.084} & \Huge0.082&\Huge16.99&\Huge3.30&\Huge0.324&\Huge35.24&\Huge1.19&\Huge0.296 \\
        \Huge RestoreFormer\cite{wang2022restoreformer}&\Huge0.830&\Huge0.842&\Huge73.88&\Huge4.86&\Huge0.707&\Huge18.59&\Huge4.53&\textcolor{blue}{\textbf{\Huge0.724}}\\
          \Huge DR2\cite{wang2023dr2}&\Huge0.740&\Huge0.753&\Huge65.03&\Huge5.00&\Huge0.619&\Huge19.09&\Huge3.98&\Huge0.560\\
          \Huge CodeFormer\cite{codeformer}  &\Huge0.826&\Huge0.849&\Huge75.27&\Huge4.92&\Huge0.712&\Huge18.64&\Huge4.56&\Huge0.670 \\
          \Huge DifFace\cite{yue2022difface}&\Huge0.727&\Huge0.762&\Huge69.01&\Huge4.56&\Huge0.661&\Huge18.58&\Huge4.07&\Huge0.570\\
          \Huge DifFace+\cite{yue2022difface} &\Huge0.795 (+0.068)&\Huge0.827 (+0.065)&\Huge73.65 (+6.64)&\Huge4.81 (+0.25)&\Huge0.696 (+0.035)&\Huge18.30 (+0.28)&\Huge4.53 (+0.46)&\Huge0.613 (+0.043)\\
        \Huge DAEFR\cite{tsai2023dual} &\Huge0.817&\Huge0.834&\Huge75.25&\Huge4.93&\Huge0.724&\Huge18.15&\Huge4.50&\Huge0.668\\
          \Huge DAEFR+\cite{tsai2023dual} &\Huge0.840 (+0.023)&\textcolor{blue}{\textbf{\Huge0.851 (+0.017)}}&\Huge75.75 (+0.50)&\Huge5.02 (+0.09)&\Huge0.730 (+0.006)&\textcolor{red}{\textbf{\Huge18.07 (+ 0.08)}}&\Huge4.59 (+0.09)&\Huge0.687 (+0.019)\\
   \Huge Interlcm\cite{li2025interlcm}&\textcolor{blue}{\textbf{\Huge0.845}}&\Huge0.849&\textcolor{blue}{\textbf{\Huge76.08}}&\textcolor{blue}{\textbf{\Huge5.08}}&\textcolor{red}{\textbf{\Huge0.753}}
          &\textcolor{blue}{\textbf{\Huge18.15}}&\textcolor{blue}{\textbf{\Huge4.75}}&\Huge0.718\\
    \textbf{\Huge Ours}&\textcolor{red}{\textbf{\Huge0.871}}&\textcolor{red}{\textbf{\Huge0.878}}&\textcolor{red}{\textbf{\Huge76.63}}&\textcolor{red}{\textbf{\Huge5.16}}&\textcolor{blue}{\textbf{\Huge0.743}}&\Huge18.16&\textcolor{red}{\textbf{\Huge4.78}}&\textcolor{red}{\textbf{\Huge0.734}}\\
      \Xhline{1.2pt}
    \end{tabular}
    }
        \caption{Quantitative comparison on the CelebA-Test dataset. 
    }
        \label{tab:synthetic}
  \end{table}

\subsection{Comparisons with State-of-the-art Methods}
\label{sec:4.3}
We compare IQPFR with state-of-the-art methods, including Interlcm~\cite{li2025interlcm}, WaveFace~\cite{miao2024waveface}, DAEFR \cite{tsai2023dual}, DR2 \cite{wang2023dr2}, DifFace \cite{yue2022difface}, CodeFormer \cite{codeformer} and RestoreFormer \cite{wang2022restoreformer}. Besides, to demonstrate the extensibility of our score-conditioned approach, we insert it into DAEFR and DifFace, notated as DAEFR+ and DifFace+.

\noindent\textbf{Evaluation on Real-world Datasets.}
IQPFR demonstrates superior performance 
across all real-world datasets. As shown in~\Cref{tab:real_world_test}, IQPFR demonstrates consistent superiority across all evaluated datasets, achieving top performance on most quality metrics. This indicates that IQPFR's restoration outputs possess superior perceptual quality relative to alternative approaches. Notably, in terms of the Musiq-GFIQA and TOPIQ-GFIQA metrics, which are specifically tailored for assessing face image quality, IQPFR significantly outperforms competing methods. Furthermore, DAEFR+ and DifFace+ outperform their baseline counterparts across all evaluation metrics, conclusively demonstrating the efficacy of of our score-conditioned approach. ~\Cref{Fig.real_world} presents qualitative comparisons between different approaches. Compared with others, IQPFR more effectively restores facial details, capturing textures and realistic colors. ~\Cref{Fig.difface+} shows that DAEFR+ and DifFace+ get better qualitative quality than their respective baseline, for example, DifFace+ demonstrates enhanced eyebrow texture and superior ocular clarity compared to its baseline counterpart in visual comparisons.


\noindent\textbf{Evaluation on Synthetic Datasets.}~\Cref{tab:synthetic} presents the quantitative comparison of various methods. IQPFR achieves state-of-the-art performance across most quantitative metrics. DAEFR+ and DifFace+ surpass DAEFR and DifFace by a large margin.
The qualitative comparison in~\Cref{Fig.celeba} further illustrates that our results are not only sharper than those of other methods but also exhibit richer textures and more aesthetically pleasing skin tones.



\paragraph{User Study.} We conduct a user study on real-world datasets by randomly selecting 60 images from three real-world datasets to compare IQPFR against DAEFR and CodeFormer.
For each image, 10 users ranked the outputs from the three methods, assigning scores of 1 to 3 points, 
\begin{wraptable}{l}{4.8cm}
\vspace{-0.3cm}
\setlength{\abovecaptionskip}{0.2cm}
\centering
  \renewcommand\arraystretch{1.1}
  \resizebox{0.65\columnwidth}{!}{
  \begin{tabular}{c|cc}
    \Xhline{1pt}
    \rowcolor{gray!30}
    & Average score&Top-1 rate \\
    \cline{1-3}
    CodeFormer&2.05&26\% \\
    DAEFR&1.52&8\% \\
    \textbf{IQPFR}&\textbf{2.42}&\textbf{75\%} \\
    \Xhline{1pt}
  \end{tabular}
  }
\caption{User study.}
\label{tab:user_study}
\vspace{-0.5cm}
\end{wraptable}
with the best result receiving 3 points and the worst receiving 1 point. We then calculated the average score for each method.
Table \ref{tab:user_study} presents the statistical results for each method. Notably, IQPFR wins the highest average score and Top-1 rate, indicating that users consistently rated IQPFR's results as the highest quality.
\begin{figure}[b]
  \setlength{\abovecaptionskip}{0cm}
    \includegraphics[width=\linewidth,scale=0.1]{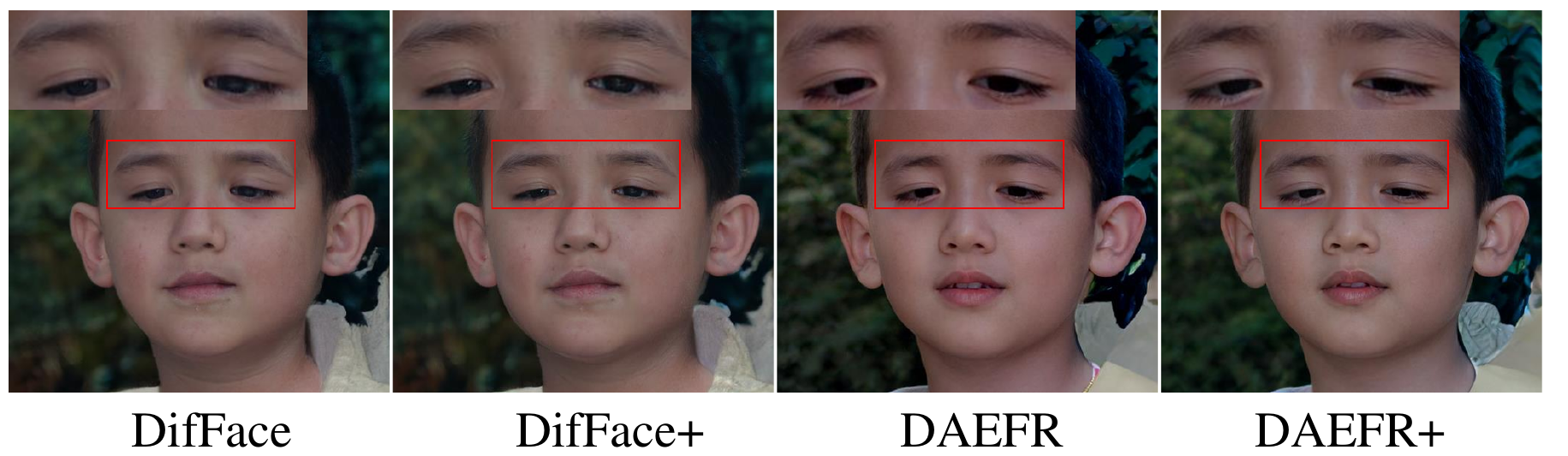}
    \caption{Effectiveness of our plug-and-play quality prior conditioned approach. DifFace+ and DAEFR+ have more textures and details than their baseline.}
    \label{Fig.difface+}
\vspace{-0.2cm}
\end{figure}
\subsection{Ablation Study}
Here, we present the ablation studies of our three algorithms on the WebPhoto-Test dataset. Additional detailed ablation studies are discussed in the appendix.


\noindent\textbf{Quality Prior Conditional Approach.}
With the quality prior conditional Transformer, we can input a target score to control the restoration quality during testing. The restoration quality improves as the score increases; for a detailed comparison, please refer to the appendix. Typically, we aim for the highest possible restoration quality, so in most situations, we input the maximum score as the condition. As shown in~\Cref{tab:ablation}, experiments (a) and (b) highlight the effectiveness of incorporating the quality prior as a condition in the codebook lookup Transformer.
~\Cref{Fig.difface+} shows the visible effectiveness of our plug-and-play quality prior conditioned approach. DAEFR+ and DifFace+ are sharper and have more details than their baseline.

\noindent\textbf{Effectiveness of Dual Codebook.}
The comparison between experiments (b) and (c) in~\Cref{tab:ablation} demonstrates that the proposed dual codebook achieves higher perceptual quality compared to the traditional single codebook approach. Visual comparisons, provided in the appendix, reveal that the result from (b) is sharper and contains richer textures than (a) and (b). The dual-codebook architecture better preserves diverse and detailed facial features, which enhances the second stage based on the codebook and decoder developed in the first stage.


\noindent\textbf{Quality Optimization.}
Experiments (c) and (d) in~\Cref{tab:ablation} demonstrate the effectiveness of quality optimization. The discrete representation provided by the codebook allows quality optimization to enhance the output quality without encountering the phenomenon of overoptimization.
\begin{table}[t]
\setlength{\abovecaptionskip}{0.2cm}
\centering

  \renewcommand\arraystretch{1.2}
  \resizebox{1\columnwidth}{!}{
  \begin{tabular}{|c|c|c|c|c|c|}
    \cline{1-6}
    ID&  \makecell[c]{Score\\ Condition}&\makecell[c]{Dual \\ Codebook}&\makecell{ Quality \\ Optimization}& TOPIQ-G$\uparrow$  & Musiq-K$\uparrow$ \\
    \cline{1-6}
    (a)& & & &0.756&73.56  \\
   (b) & $\surd$ & &  &0.791&75.96 \\
   (c) & $\surd$ & $\surd$ &&0.805 &76.02  \\
   (d)  & $\surd$ & $\surd$ &$\surd$ &\textbf{0.820}&\textbf{76.80}  \\
    \cline{1-6}
  \end{tabular}
  }
  \caption{
Ablation study of quality prior.}
  \label{tab:ablation}
\vspace{-0.8cm}

\end{table}
\begin{table}[b]
\setlength{\abovecaptionskip}{0.2cm}
\centering
\renewcommand\arraystretch{1.1}
\resizebox{1\columnwidth}{!}{
  \begin{tabular}{c|cc}
    \Xhline{1pt}
    \rowcolor{gray!30}
    & \textbf{Continuous}& \textbf{Discrete}\\
     \Xhline{1pt}
    \makebox[0.05\textwidth][c]{\makecell[c]{Average\\ Quality}}&0.93&0.87 \\
    \Xhline{1pt}
    Example&  \makecell[c]{\includegraphics[scale=0.22]{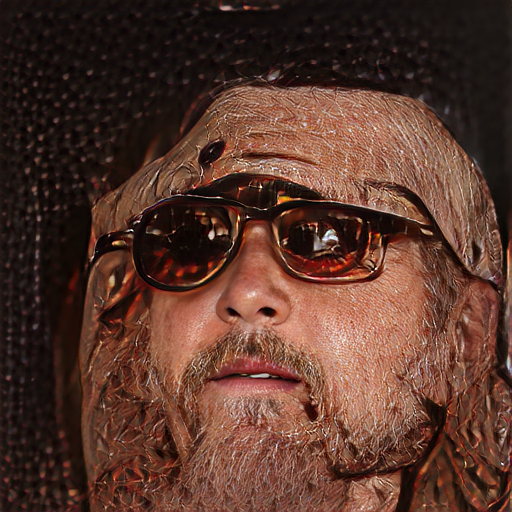}}&  \makecell[c]{\includegraphics[scale=0.22]{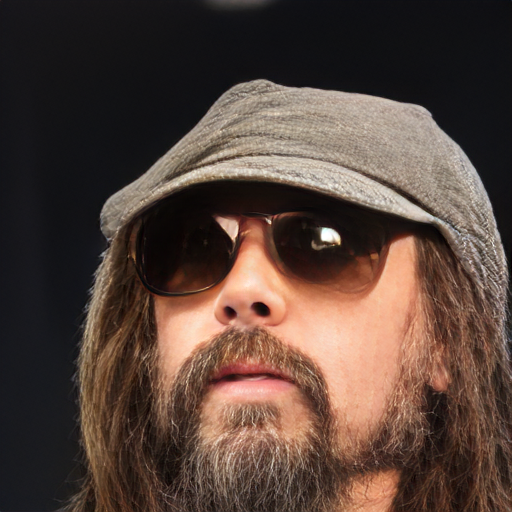}}\\
   \Xhline{1pt}
  \end{tabular}
  }
\caption{Quality optimization in continuous and discrete representation space.}
  \label{tab:overopti}
\end{table} 
\noindent\textbf{Discrete Representation Mitigate Overoptimization.}
\Cref{tab:overopti} shows the results of using quality optimization in continuous representation space and discrete representation space. In continuous representation space, the results get high-quality scores but low human perception. However, in discrete representation space, the results get high-quality scores while keeping high human perception at the same time, which means the discrete representation space is able to mitigate the over-optimization problem.

\subsection{Extended Applications}
To further illustrate the extensibility of our plug-and-play quality prior conditioned approach. We conduct experiments on two other tasks, Face Color Enhancement and Underwater Image Enhancement.

\begin{table}[t]
\setlength{\abovecaptionskip}{0.2cm}
\centering

  \renewcommand\arraystretch{1.1}
  \resizebox{1\columnwidth}{!}{
  \begin{tabular}{c|ccccc}
    \Xhline{1pt}
    \rowcolor{gray!30}
    Methods& PSNR$\uparrow$&SSIM$\uparrow$&Uranker\cite{guo2023underwater}$\uparrow$& UCIEQ\cite{yang2015underwater} $\uparrow$& UIQM\cite{panetta2015human}$\uparrow$ \\
    \cline{1-6}
    PUIE-Net&20.55&0.8418&1.8247&0.5666&\textbf{2.988} \\
    PUIE-Net+&\textbf{20.60}&\textbf{0.8430}&\textbf{1.9160}&\textbf{0.5734}&2.974 \\
    \cline{1-6}
    NU2Net&\textbf{22.49} &0.9248 &1.8265 &0.5935 & 2.757 \\
   NU2Net+&22.47 &\textbf{0.9259} & \textbf{1.9306} &\textbf{0.6038} &\textbf{2.812}  \\
    \cline{1-6}
  \end{tabular}
  }
\caption{Results on the underwater image enhancement task.}
  \label{tab:underwater}
\end{table}

\begin{figure}[t]
  \setlength{\abovecaptionskip}{0cm}
  \captionsetup[subfigure]{font=small,labelformat=empty,justification=centering}
  \centering
  \begin{subfigure}[t]{0.24\columnwidth}
      \includegraphics[scale=0.12]{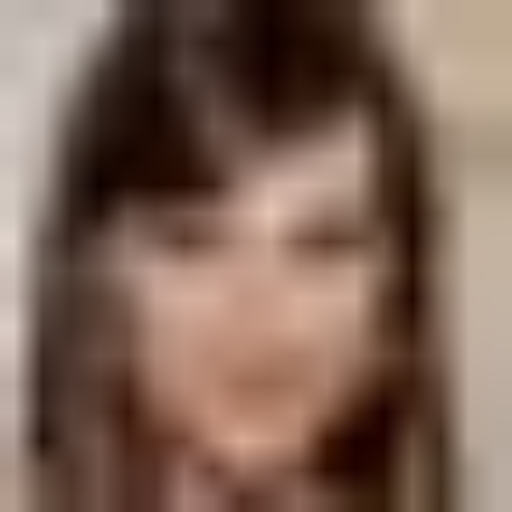}\subcaption{LQ}
  \end{subfigure}
  \begin{subfigure}[t]{0.24\columnwidth}
      \includegraphics[scale=0.12]{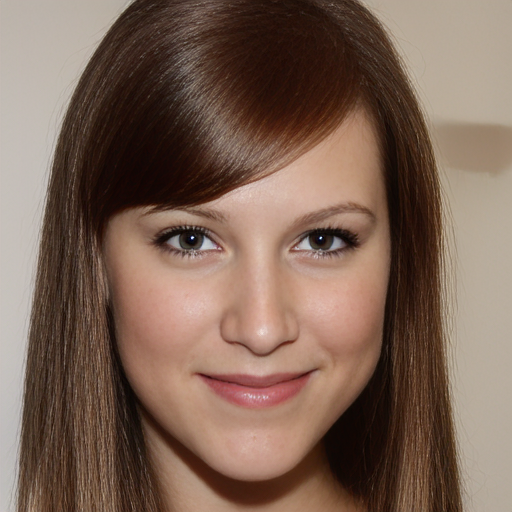}\subcaption{CodeFormer}
  \end{subfigure}
    \begin{subfigure}[t]{0.24\columnwidth}
      \includegraphics[scale=0.12]{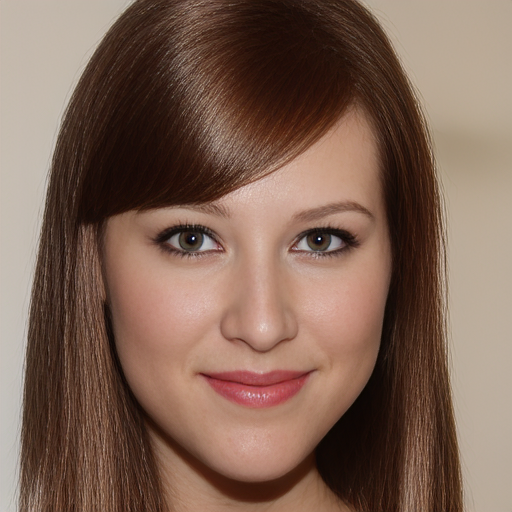}\subcaption{CodeFormer+}
  \end{subfigure}
  \begin{subfigure}[t]{0.24\columnwidth}
      \includegraphics[scale=0.12]{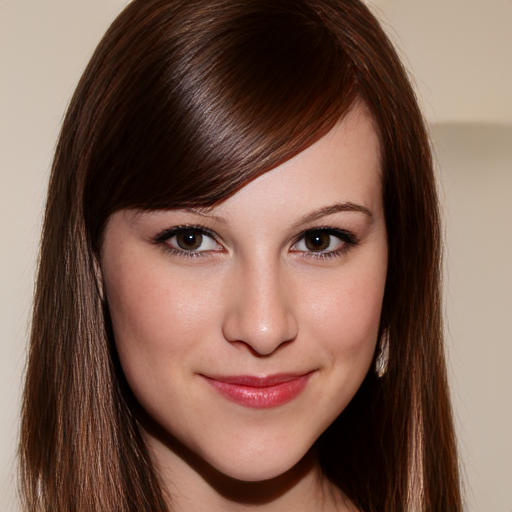}\subcaption{Ours}
  \end{subfigure}
  \caption{Qualitative comparison on Face Color Enhancement.}
  \label{Fig.facecolor}\vspace{-2mm}
\end{figure}

\noindent\textbf{Face Color Enhancement.} We compare IQPFR with codeformer and codeformer+. In~\Cref{Fig.facecolor} our results show better color aesthetics and more facial textures than others. And Codeformer+ has more natural color than Codeformer.

\begin{figure}[ht]
  \setlength{\abovecaptionskip}{0cm}
  \captionsetup[subfigure]{font=small,labelformat=empty,justification=centering}
  \centering
  \begin{subfigure}[t]{0.24\columnwidth}
  \includegraphics[scale=0.057]{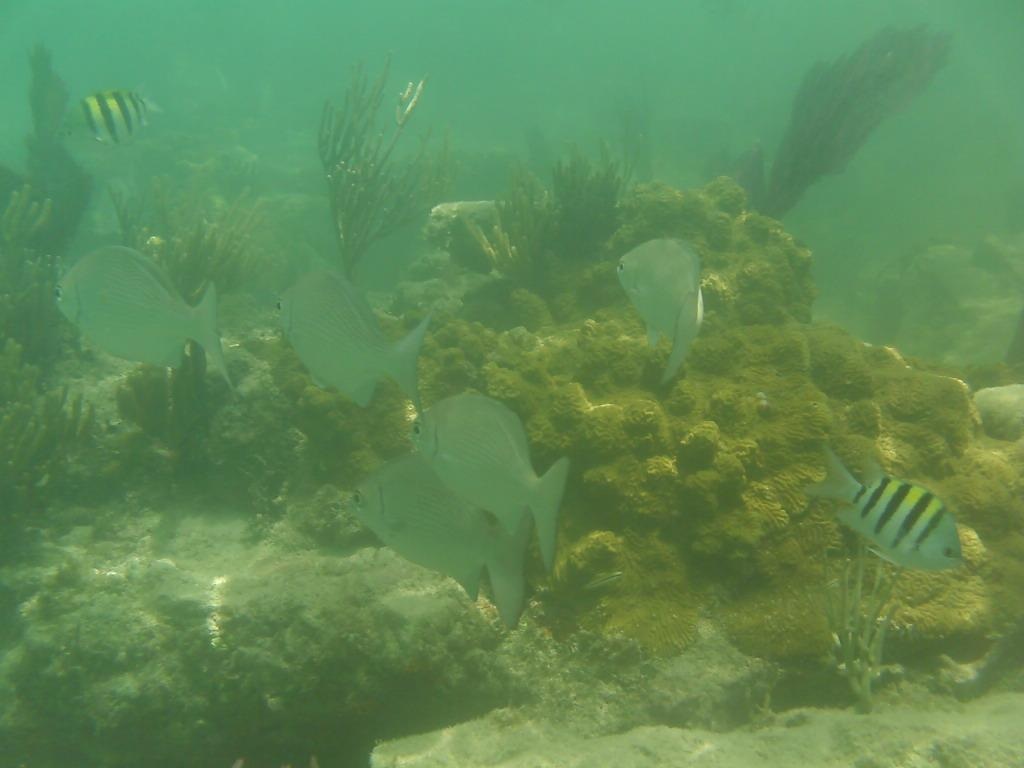}\subcaption{LQ}
  \end{subfigure}
  \begin{subfigure}[t]{0.24\columnwidth}
      \includegraphics[scale=0.057]{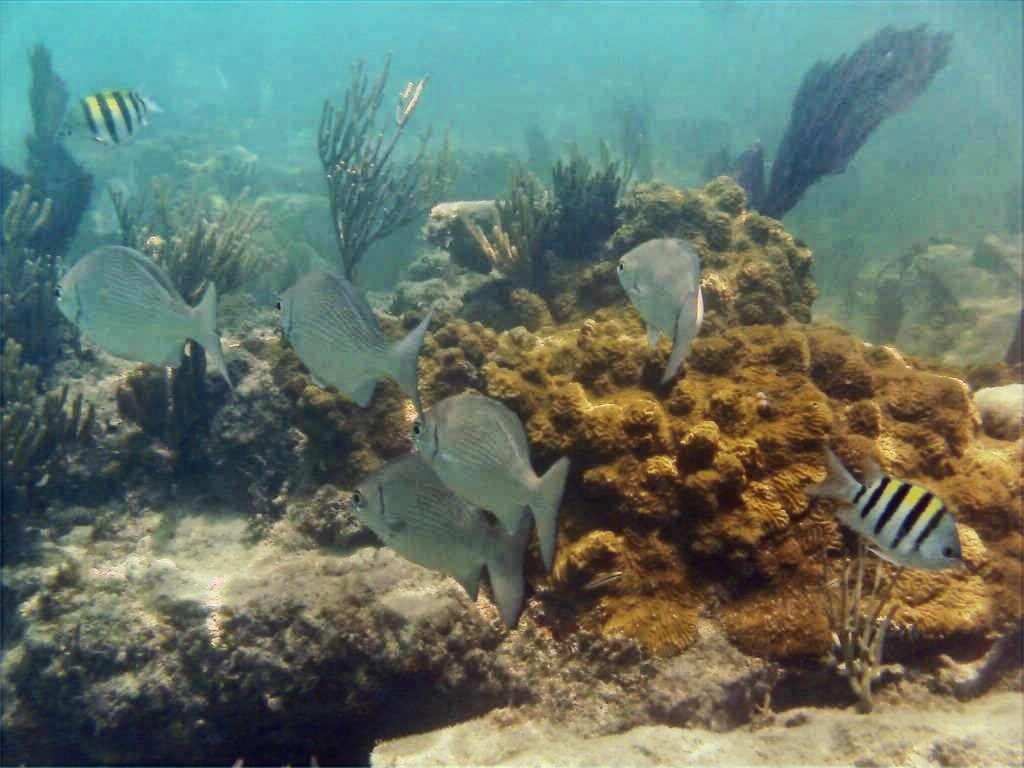}\subcaption{NU2Net}
  \end{subfigure}
  \begin{subfigure}[t]{0.24\columnwidth}
      \includegraphics[scale=0.057]{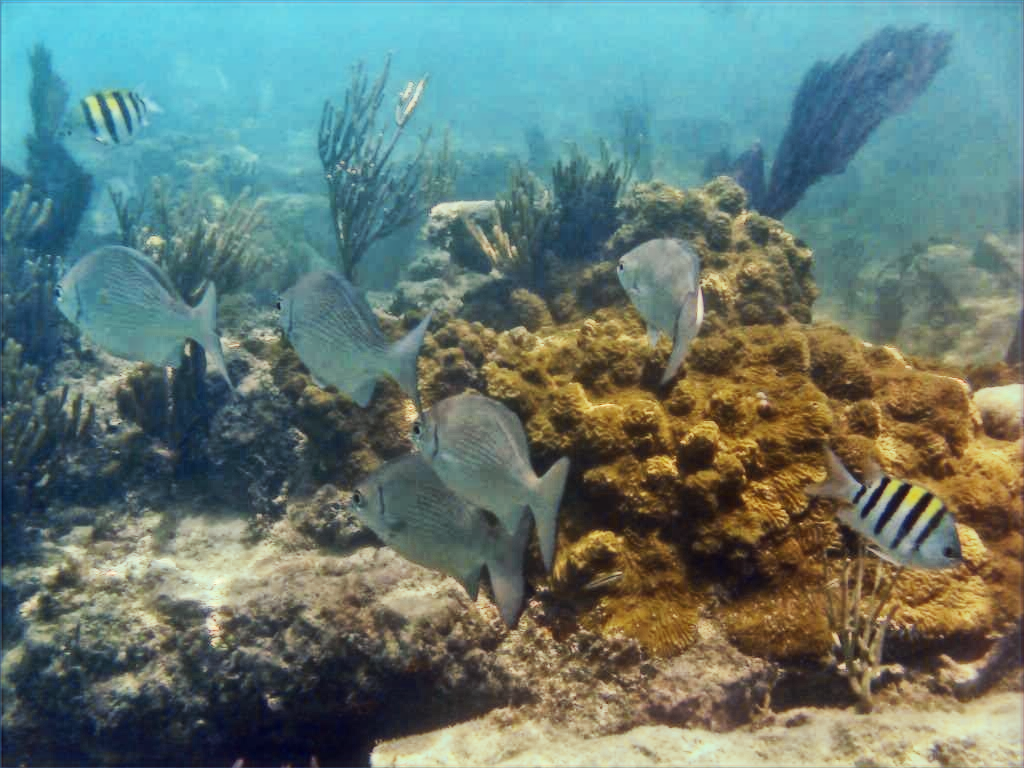}\subcaption{NU2Net+}
  \end{subfigure}
  \begin{subfigure}[t]{0.24\columnwidth}
      \includegraphics[scale=0.057]{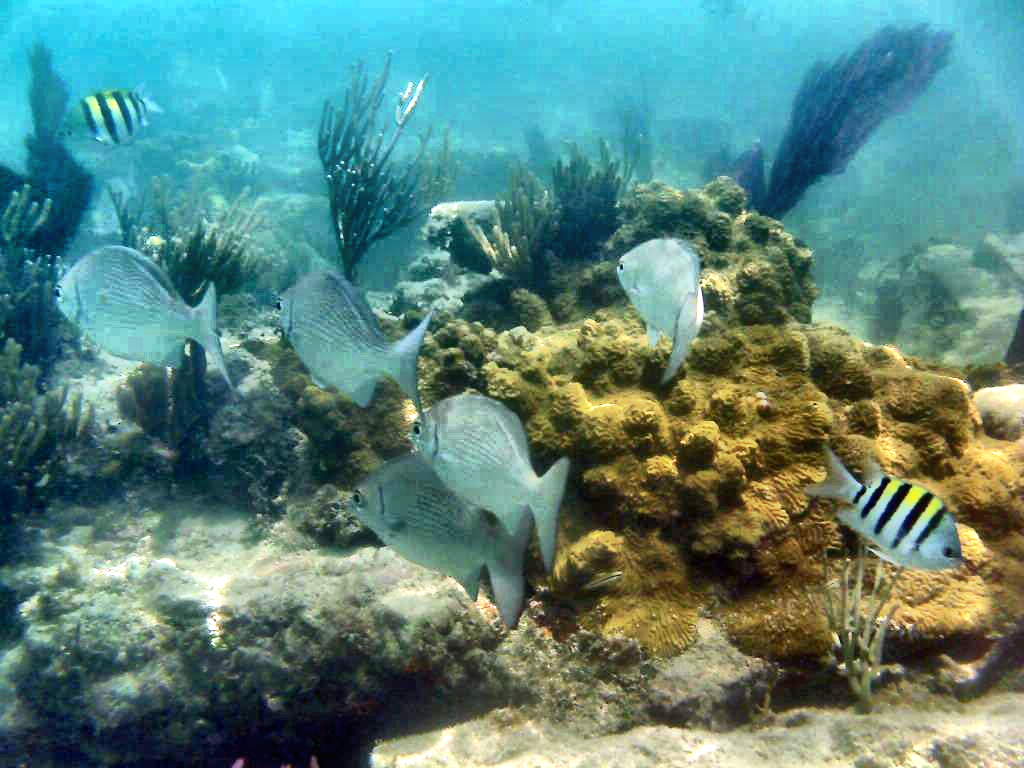}\subcaption{HQ}
  \end{subfigure}
  \caption{Results on Underwater Image Enhancement task.}
  \label{Fig.underwater}\vspace{-5mm}
\end{figure}

\noindent\textbf{Underwater Image Enhancement.}
For the underwater image enhancement (UIE) task, we take PUIE-Net~\cite{fu2022uncertainty} and NU2Net~\cite{guo2023underwater} as baseline models. We integrate score conditions into both models to obtain PUIE-Net+ and NU2Net+. And we take some widely used metrics in the UIE task to evaluate the performance. Table \ref{tab:underwater} and Figure \ref{Fig.underwater} show the quantitative and qualitative comparisons. The score conditional approach brings a large performance increment to both models. And NU2Net+ shows better ability in correcting underwater color aberrations.
\vspace{-0.15cm}
\section{Conclusion}
\vspace{-0.15cm}
In this paper, we propose IQPFR to incorporate quality priors into BFR  for high-quality content generation. 
First, we introduce a dual-codebook autoencoder, which consists of a common codebook and an HQ\textbf{+} codebook to learn diverse and high-quality facial features. 
We then propose a plug-and-play score-conditioned 
Transformer during code prediction. Finally, we perform quality optimization based on the discrete representation space, which mitigates the over-optimization problem. 
Our experiments demonstrate our superiority in performance and generalization.

%% file: hp_sec/X_suppl.tex
\begin{figure*}[ht]
  \captionsetup[subfigure]{font=scriptsize,labelformat=empty,justification=centering}
  \centering
  \includegraphics[scale=0.35]{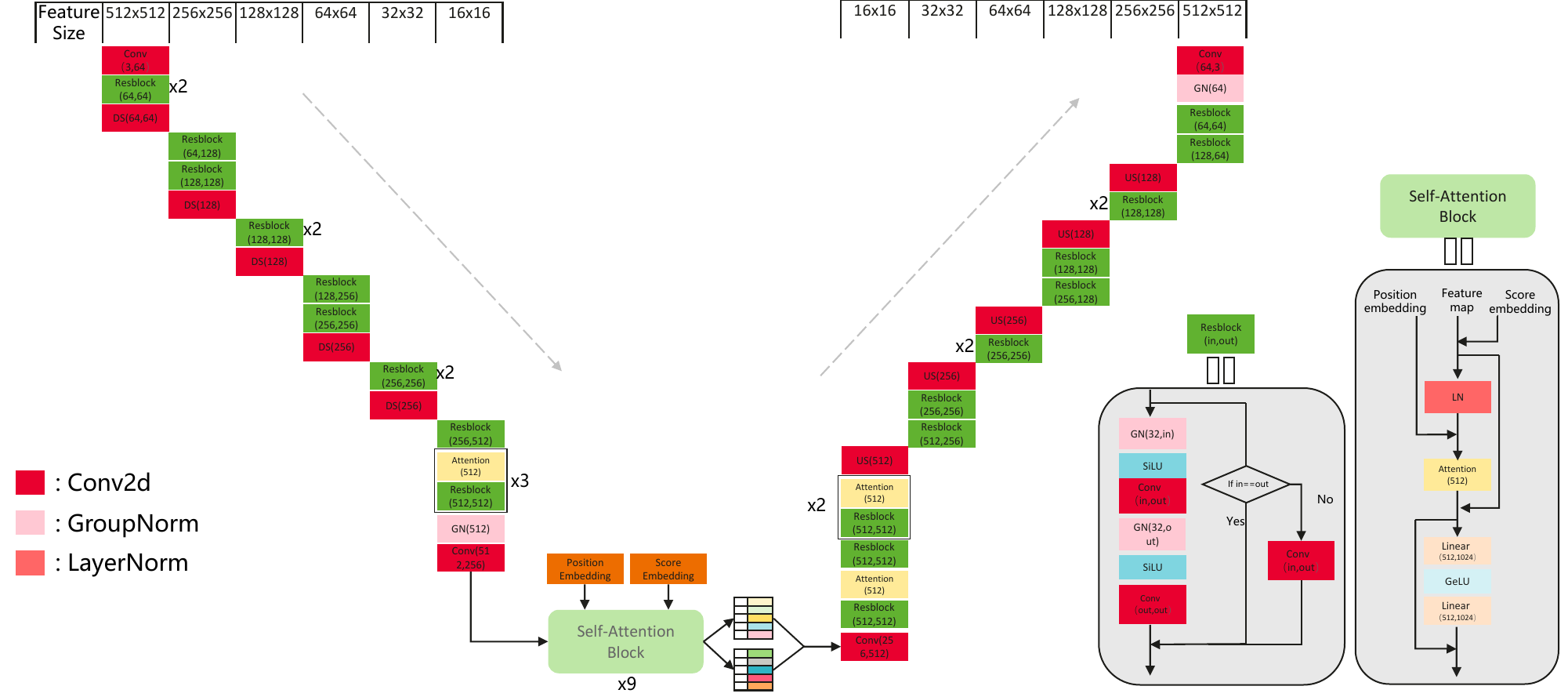}.
\caption{The detailed architecture of Stage II model.}
\label{Fig.detailed_architecture}
\end{figure*}

\newpage
\mbox{}
\newpage

\section{Appendix}
We present additional content in this section to complement our main manuscript. 
\Cref{sec:6.1} provides further details about our model architecture and experimental settings. 
\Cref{sec:6.2} includes additional ablation studies to elucidate our proposed method. 
\Cref{sec:limitation} discuss the limitations of our work.
\Cref{sec:6.4} discuss the possible social impacts of our work.
Finally, \Cref{sec:6.5} showcases more qualitative comparisons across all testing datasets.
\begin{figure*}[ht]
  \captionsetup[subfigure]{font=scriptsize,labelformat=empty,justification=centering}
  \centering
  \includegraphics[scale=0.35]{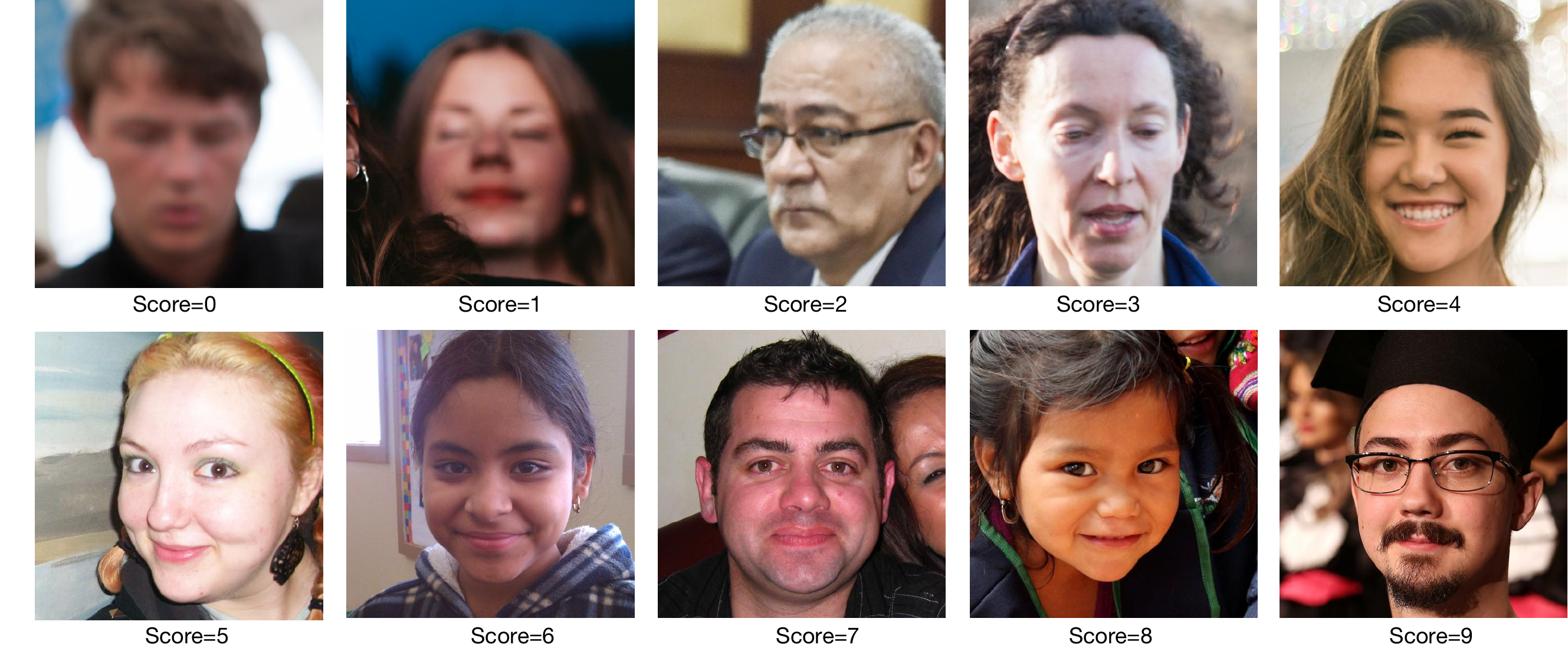}.
\caption{Examples of pictures at different quality scores. The score is an integer between 0 and 9. As the score increases, the image quality gets better with more textures and details.}
\label{Fig.ffhq_example}
\end{figure*}
\begin{table*}[ht]
  \caption{Ablation study of the size of HQ\textbf{+} codebook and the fusion methods of quantized features.
  }
  \label{tab:abl_codebooksize}
  \centering
  \renewcommand\arraystretch{1.2}
  \scalebox{1}{
  \begin{tabular}{ccccc}
    \Cline{1-5}
    Methods & TOPIQ-G$\uparrow$& Musiq-G$\uparrow$& Musiq-K$\uparrow$  & Musiq-A$\uparrow$\\
    \cline{1-5}
    256 + Addition&0.847& 0.861&\textcolor{blue}{75.42} &5.12\\
    512 + Addition&\textcolor{blue}{0.850}&\textcolor{red}{0.862} &75.37 &\textcolor{blue}{5.16} \\
    1024 + Addition&\textcolor{red}{0.853}&\textcolor{blue}{0.861} &\textcolor{red}{75.53} &\textcolor{red}{5.18} \\
    1024 + Attention&0.831&0.843&75.34 &5.02 \\
    \Cline{1-5}
  \end{tabular}
  }
\end{table*}
\subsection{Implementation Details.}
\label{sec:6.1}
\subsubsection{Model Architecture.}The detailed architecture of our second-stage model is illustrated in \Cref{Fig.detailed_architecture}. Our model can be divided into three parts: the LQ Encoder, the Conditional Transformer and the HQ Decoder.
The LQ Encoder mainly consists of residual blocks and convolution operators, effectively encoding LQ image $x_l\in \mathbb{R}^{512\times512\times3} $ into a feature map $Z_{l}\in \mathbb{R}^{16\times16\times256}$.
The Conditional Transformer comprises 9 Self-Attention layers. The first layer takes $Z_{l}$ and the condition score as inputs, while subsequent layers use the output of the preceding layer as their input.
The Transformer outputs two code index sequences, $\textbf{c}_1$ and $\textbf{c}_2$. These sequences then receive code entries from the common codebook and the HQ\textbf{+} codebook, respectively, forming quantized features $Z_f^1$ and $Z_f^2$. The two quantized features are subsequently combined as described in \Cref{eq:z1_z2} of the main manuscript to obtain $Z_f$. The Decoder is also primarily composed of residual blocks and convolution operators. $Z_f$ is fed to the Decoder and then decoded back into a high-quality image, $x_{res}$.

\subsubsection{Experimental Setting.}
We set the balance weight of quantized feature fusion $\alpha$ in \Cref{eq:dual_codebook} to 1.0.
The loss weight $\beta$ in \Cref{eq:loss_lfeat} is set to 0.25, and $\lambda_1$ and $\lambda_2$ in \Cref{eq:loss_stage2} are 0.5 and 0.1 respectively.
For all three quality priors, we initially normalize the score of FFHQ dataset as follows:
\begin{equation}
  \bar{s}_i=\frac{s_i-s_{min}}{s_{max}-s_{min}+0.00001}, i \in [0,70000).
\end{equation}
The score threshold $S_{thr}$ in ~\Cref{eq:dual_codebook} determines the number of HQ\textbf{+} images.
A small $S_{thr}$ leads to a small quality gap between the HQ\textbf{+} codebook and the common codebook, 
while a large $S_{thr}$ makes it more challenging to learn the HQ\textbf{+} well.
We set $S_{thr}$ of the mean score in \Cref{eq:dual_codebook} to 0.90.
With this configuration, the number of HQ\textbf{+} images is approximately 8k, a balance we found suitable for the trade-off between quality gap and learning complexity according to our experiments.
During training, we multiply the score by 10 and round it down, ensuring the condition score falls within an integer range of [0,9].

As stated in the main manuscript (line 305-306), with quality score conditional injection, we can control the quality of output results during the testing phase. However, it is important to note that the controllability itself is not our ultimate objective – our final goal is to leverage this controllability to achieve the highest possible quality output.
So during testing, the condition scores of our quantitative and qualitative results in \Cref{tab:real_world_test}, \Cref{tab:synthetic},  \Cref{Fig.real_world} and \Cref{Fig.celeba}  of main manuscript are all 9.

\subsubsection{Training Data Degradation/}
For consistency and fair comparison, we adopt the same degradation method used in previous studies\cite{codeformer,tsai2023dual}:
We train our model on the FFHQ dataset \cite{karras2019style}, which contains 70k HQ face images.
All images are resized to $512\times 512$. During the training of the second stage, we manually degrade the HQ image $I_h$ to generate the HQ-LQ pair.
And for a fair comparison, we use the same degradation method as previous work \cite{codeformer,tsai2023dual,wang2023dr2}:
\begin{equation}
  \centering
  \begin{aligned}
        I_l=\{[(I_h\otimes k_\sigma){\downarrow_r} +n_\delta]_{JEPG_q}\}{\uparrow_r}.
    \label{eq:degrade}
  \end{aligned}
\end{equation}
First, $I_h$ is blurred by a Gaussian kernel $k_\sigma$ via convolution operator $\otimes$. Then we down-sample the image with a scaling factor $r$.
Subsequently, a Gaussian noise $n_\delta$ is added. And we apply a JPEG compression with a quality factor $q$. Finally, the degraded image is upgrading back to a resolution of $512\times512$ to obtain the final degraded image $I_l$.
$\sigma$, $r$, $\delta$, and $q$ are randomly sampled from [1, 15], [1, 30], [0, 20], and [30, 90] respectively.

\subsection{More Ablation Studies.}
\subsubsection{The analysis of FFHQ dataset.}
As show in \Cref{Fig.ffhqscore}, not all images in FFHQ are of sufficiently high quality. The average score of FFHQ is around 0.8, which means that the recovery quality of the traditional restoration model trained on this dataset is also 0.8, and it is usually difficult to achieve this ideal situation due to the limitation of training.
\Cref{Fig.ffhq_example} shows an example of different quality images in FFHQ, those images with degradation limit the overall restoration quality. 
Our image quality prior achieves to improve the average quality of the restoration results to the highest quality of FFHQ without altering the dataset by guiding the training of the model to differentiate between different quality GT images.
It is important to note that our methods differ from approaches that use NR-IQA models for data filtering, where only HQ\textbf{+} images are used for training. Since the highest-quality images are often in the minority, they may lack sufficient diversity in facial information. Relying solely on these images for training could lead to facial defects in the outputs (e.g., distorted teeth). In contrast, our common codebook contains rich and diverse facial features, ensuring the model's robustness across various input types.

\label{sec:6.2}
\begin{figure*}[ht]
  \captionsetup[subfigure]{font=scriptsize,labelformat=empty,justification=centering}
  \centering
  \includegraphics[scale=0.25]{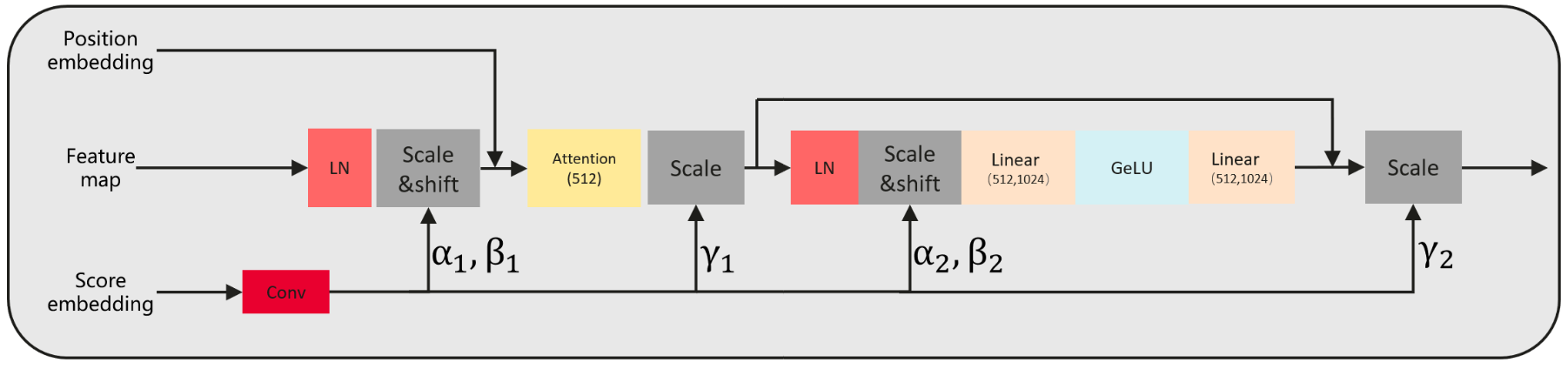}.
\caption{The architecture of AdaIN.}
\label{Fig.adain}
\end{figure*}
\begin{table*}[ht]
  \caption{Quantitative comparison on the real world datasets. $\uparrow$ means higher is better and $\downarrow$ means lower is better.
  }
  \label{tab:injection_method}
  \centering
  \renewcommand\arraystretch{1.2}
  \scalebox{1}{
  \begin{tabular}{|c|c|c|c|c|c|c|}
    \cline{1-7}
    Datasets&Methods & TOPIQ-G$\uparrow$& Musiq-G$\uparrow$ & Musiq-K$\uparrow$  & Musiq-A$\uparrow$& FID$\downarrow$ \\
    \cline{1-7}
  \multirow{2}{*}{LFW-Test}
          &AdaIN  &0.845&0.858&\textbf{76.20}&4.92 &  56.24         \\
          &Add    &\textbf{0.857}&\textbf{0.867} &76.19&\textbf{4.96} &\textbf{53.84}          \\
  \cline{1-7}
  \multirow{2}{*}{WebPhoto-Test}
          &AdaIN   &0.821&0.841&75.87&4.96&\textbf{88.12}           \\
          &Add     &\textbf{0.835}&\textbf{0.857}  &\textbf{76.03}&\textbf{5.03}&89.46       \\
  \cline{1-7}
  \multirow{2}{*}{WIDER-Test}
          &AdaIN    &0.831&0.856   &74.99&4.78& 39.84      \\
          &Add      &\textbf{0.845}&\textbf{0.868}  &\textbf{75.01}&\textbf{4.83}&\textbf{38.74}      \\
  \cline{1-7}
  \end{tabular}
  }
\end{table*}
\subsubsection{The Effect of HQ\textbf{+} Codebook Size.}

As the number of HQ\textbf{+} images is fewer than that of common images, we consider using an HQ\textbf{+} Codebook with a smaller size than the common codebook.
We conducted ablation studies to evaluate the effect of varying the size N of HQ\textbf{+} codebook on face image reconstruction.
\Cref{tab:abl_codebooksize} presents the quantitative reconstruction comparison on the CelebA-Test. 
A smaller HQ\textbf{+} codebook size leads to minor performance degradation, so we selected 1024 as the size for the HQ\textbf{+} codebook.
\subsubsection{Different Fusion Methods of Quantized Feature.}
We explore different methods for fusing quantized features, including the attention mechanism and direct addition. 
In the case of the attention mechanism, when $S > S_{thr}$, we implement cross-attention between the common codebook quantized feature $Z^1_f$ and the HQ\textbf{+} codebook quantized feature $Z^2_f$.
 Here, $Z^2_f$ is used to generate the key \textbf{K} and value \textbf{V}, while $Z^1_f$ is employed to generate the query \textbf{Q}. 
 If $S \leq S_{thr}$, we apply self-attention to $Z^1_f$:
\begin{equation}
    \textbf{Q}=Z_f^1\textbf{W}_q+\textbf{b}_q, \\
\end{equation}
\begin{equation}
  \begin{cases}
\textbf{K}=Z_f^2\textbf{W}_k+\textbf{b}_k,  if \text{ } S>S_{thr} ,  \\
\textbf{K}=Z_f^1\textbf{W}_k+\textbf{b}_k,  if \text{ } S <= S_{thr}.\\
  \end{cases}
\end{equation}
\begin{equation}
\begin{cases}
\textbf{V}=Z_f^2\textbf{W}_v+\textbf{b}_v&, if \text{ } S>S_{thr} ,  \\
\textbf{V}=Z_f^1\textbf{W}_v+\textbf{b}_v&, if \text{ } S <= S_{thr}.\\
  \end{cases}
  \label{eq:dual_codebook_attention}
\end{equation}
And when using attention mechanism, \Cref{eq:dual_codebook} is rewrited to:
\begin{equation}
Z_q=MultiHeadAttention(Q,K,V).
\end{equation}
\Cref{tab:abl_codebooksize} provides a quantitative reconstruction performance comparison on the CelebA-Test dataset, demonstrating that the attention mechanism fails to enhance performance despite increasing the parameter count. Therefore, we opted to directly add the quantized features in our experiments.

\subsubsection{Different Methods of Incorporating Condition.}
We incorporate quality score into the second stage of training, here we discuss two different methods of condition injection: (a) Add to $Z_l$ after embedding. (b) Adaptive layer norm.
\Cref{3.2} in main manuscript shows the detailed process of (a). And for (b), we replace standard normalization layers in transformer blocks with adaptive layer norm (AdaIN) \cite{perez2018film}.
The process of AdaIN is shown in \Cref{Fig.adain}. And \Cref{tab:injection_method} shows the quantitative comparison of two methods on real-world datasets.
While AdaIN increases the number of model parameters, it unfortunately results in inferior performance.

\subsubsection{Visualization of Ablation Study in Main Paper.}
\begin{figure*}[t]
    \includegraphics[scale=0.5]{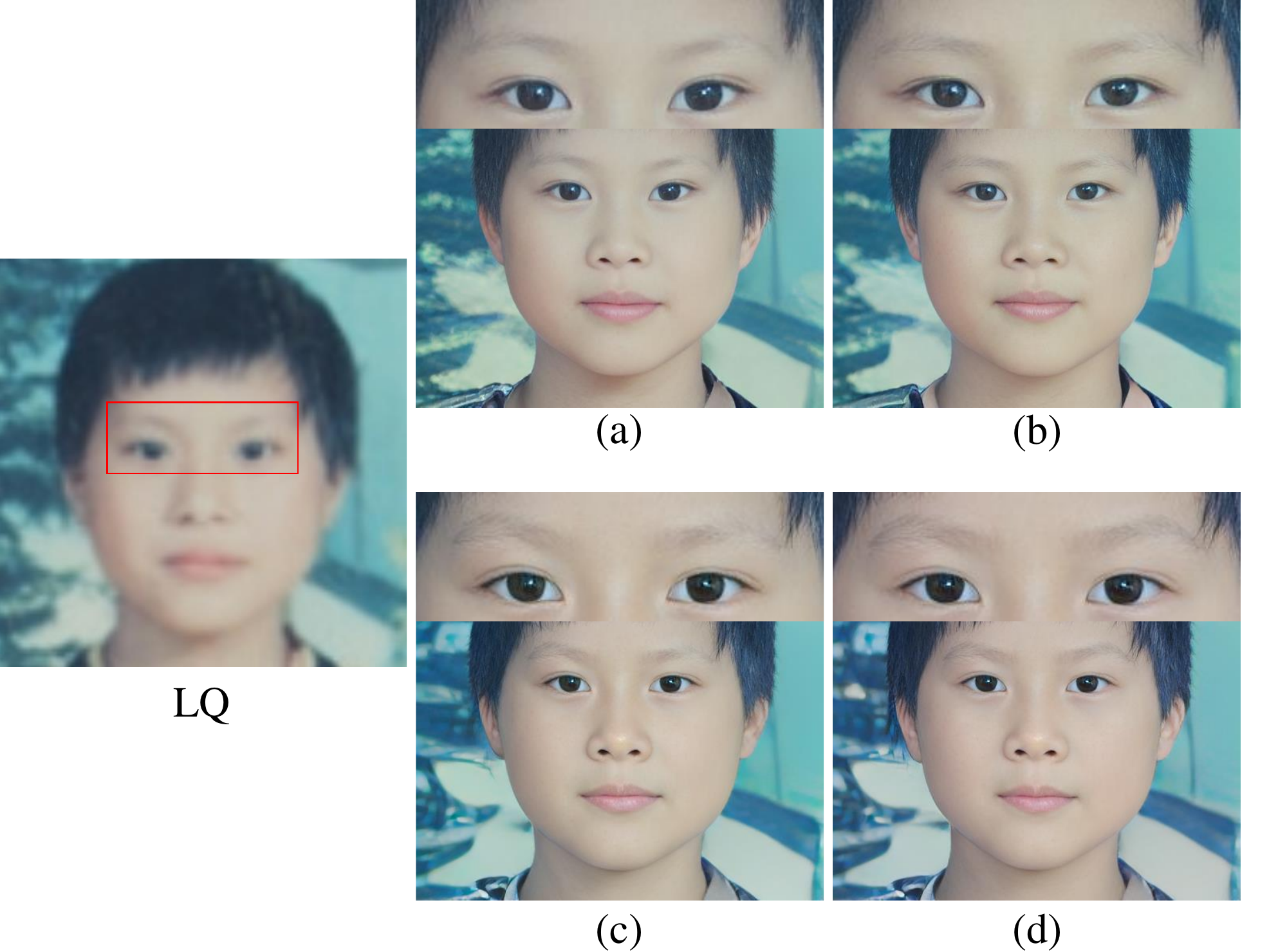}
\caption{Visualization of ablation study. (a), (b), (c) and (d) is consistent with ~\Cref{tab:ablation}. (a) is the baseline model, (b) is the baseline model with our score conditioned approach, (c) incorporates our dual codebook based on (b), and (d) is our final model with three algorithms. (b) has more details than (a). (c) has more details than (b). (d) has more texture and better color aesthetic than (c). Zoom in to see the details.}
\vspace{-0.6cm}
\label{Fig.ablation_vis}
\end{figure*}

As shown in \Cref{Fig.ablation_vis}, The result of (b) is sharper and has richer textures than (a), which demonstrates the HQ\textbf{+} codebook is able to preserve detailed facial features and thereby improves the restoration quality. Experiments (b) and (c) demonstrate the effectiveness of incorporating quality prior as condition of codebook lookup transformer. The result of (b) is sharper and has mode details than (c). And experiments (c) and (d) show the effectiveness of quality optimization. Benefiting from the discrete representation of codebook, quality optimization can increase the outputs' quality without the phenomenon of overoptimization.

\subsubsection{The Effect of Condition Score during Testing.}
\begin{figure*}[t]
  \captionsetup[subfigure]{font=tiny,labelformat=empty,justification=centering}
  \centering
  \includegraphics[scale=0.5]{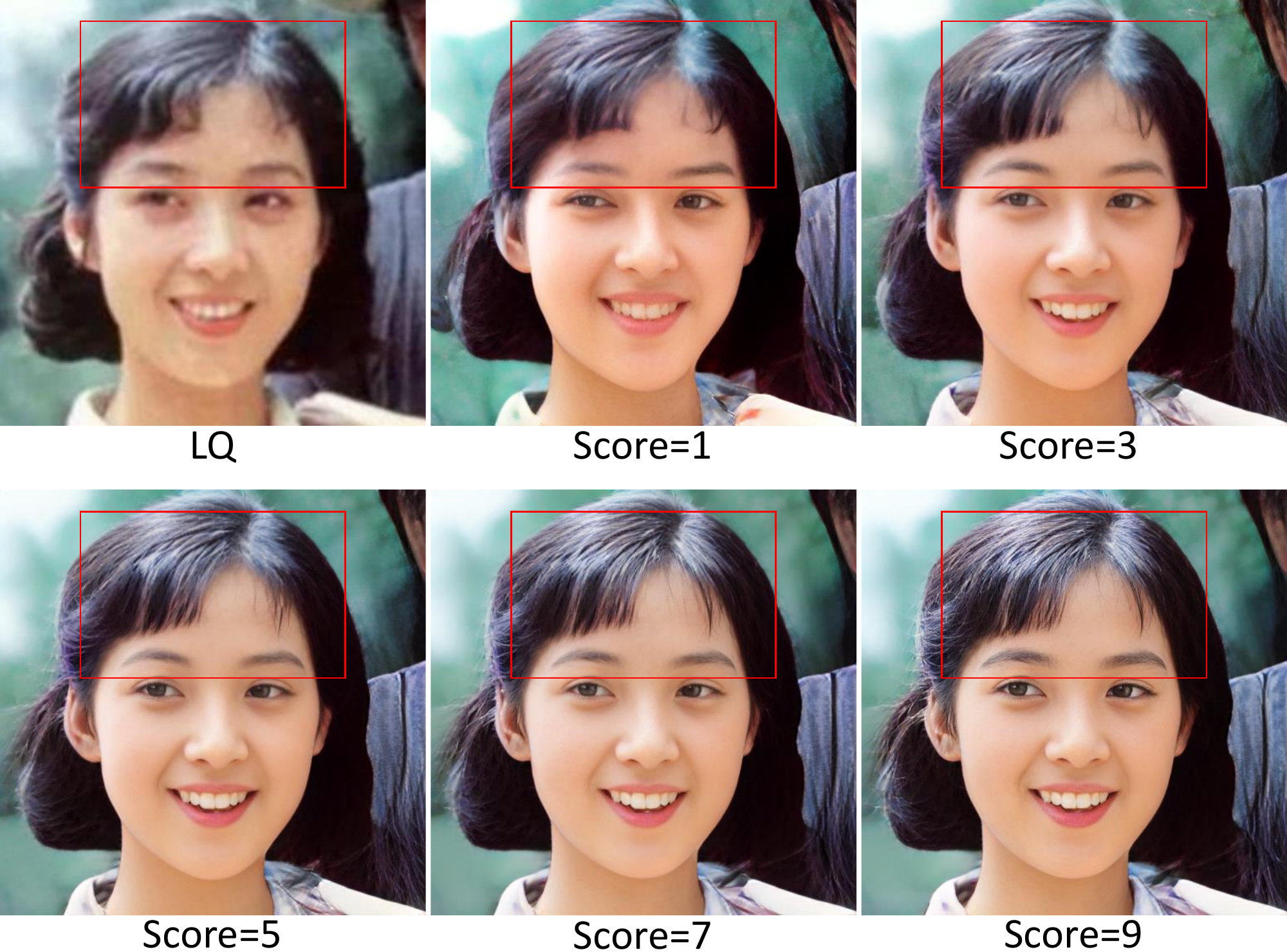}
  \caption{Inputting different target score as condition leads to different restoration quality. The higher target score is, the higher restoration quality is.   Zoom in to see the details.}
  \label{Fig.score_control}
\end{figure*}

We demonstrate that the condition score can significantly impact restoration quality during testing. 
As illustrated in \Cref{Fig.score_control}, the restoration quality can be manipulated by inputting different target scores as conditions. 
Typically, our goal is to achieve the highest restoration quality, leading us to input the highest score as the condition. 
Readers might think that for BFR task, it is not very important to incorporate quality control capability because we always want the quality to be as high as possible.
It needs to be emphasized that the purpose of using scores as conditions is that when we input the highest score, the output quality is the highest. In here the quality control capability is merely a method to achieve the highest output quality; the control capability itself is not our ultimate goal.

\subsection{Limitation.}
\label{sec:limitation}
\vspace{-0.15cm}
IQPFR combines quality priors and codebook priors to achieve high quality generation and high fidelity preservation. 
There are two main limitations resulting from quality prior and codebook prior respectively: 
(a) We use existing NR-IQA model to obtain the quality prior for IQPFR, however, the image quality assessed by IQA models is not perfectly accurate.
The limitation of IQA models impacts the effectiveness of our method. We believe our method can be further benefited as the development IQA models.
(b) The finite output space resulted from discrete codebook prior limits the diversity of outputs. 
Some facial features and accessories which are rarely existing in training data cannot be decoded perfectly. How to efficiently utilize the limit codebook is also our future work.

\subsection{Discussion of Societal Impacts.}
\label{sec:6.4}
Since our task focus on face images, there are some positive and negative societal impacts of our work.
Regrading positive impacts, our work is able to restore LQ face images with unknown degradation to HQ face images, which is widely demanded in daily life. For example, people would want to restore their valuable images which are degraded due to all kinds of reasons. As for negative impacts, no methods can preserve the identity of subjects perfectly due to the ill-posed nature of BFR task, including our methods. The change of identity may bring some negative impacts. For example, if people want to recognize the subject according to the restored face image, the change of identity will mislead them.

\subsection{More Visual Comparison.}
\begin{figure*}[t]
  \setlength{\abovecaptionskip}{-0.2cm}
  \captionsetup[subfigure]{font=scriptsize,labelformat=empty,justification=centering}
  \centering
  \includegraphics[scale=0.48]{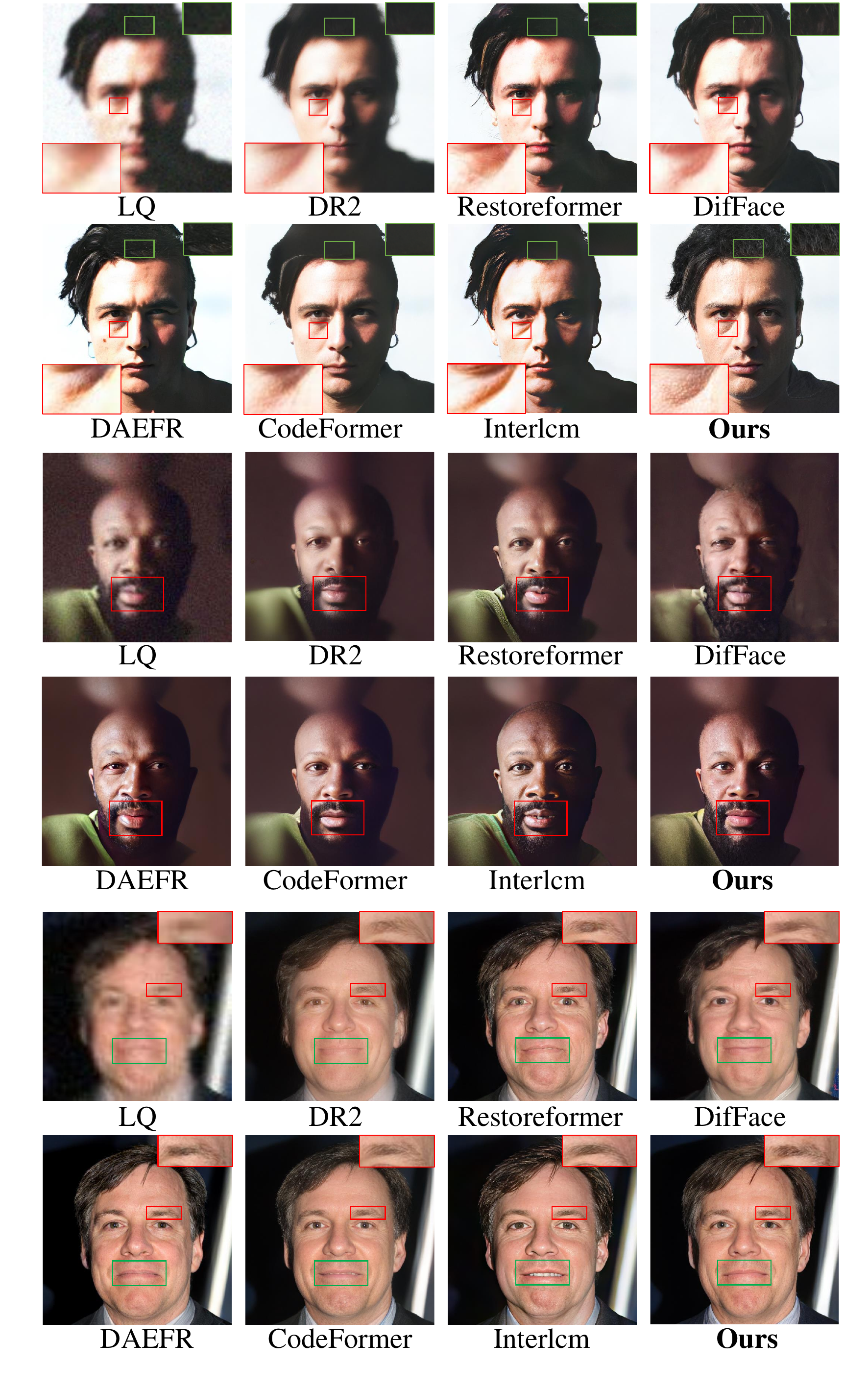}
  \caption{The qualitative comparisons on the CelebA-Test. }
  \label{Fig.vis_celeba}
\end{figure*}

\begin{figure*}[t]
  \setlength{\abovecaptionskip}{-0.2cm}
  \centering
  \captionsetup[subfigure]{font=scriptsize,labelformat=empty,justification=centering}
  \includegraphics[scale=0.48]{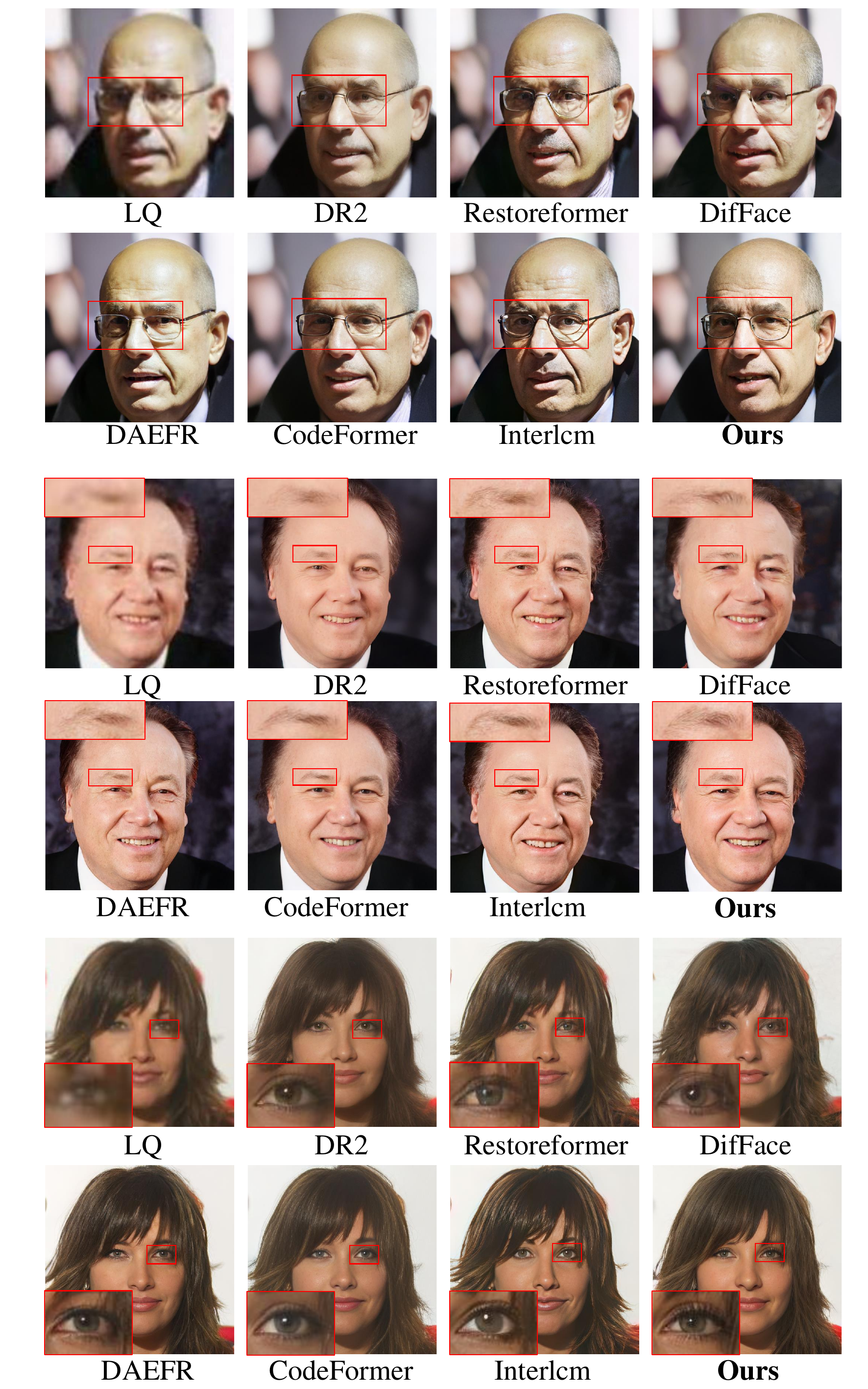}
  \caption{The qualitative comparisons on the LFW-Test. Our results are sharper and have more facial details (e.g., eyebrows and eyelashes).  Please zoom in for the best view.}
  \label{Fig.vis_lfw}
\end{figure*}

\begin{figure*}[t]
  \setlength{\abovecaptionskip}{-0.4cm}
  \centering
  \captionsetup[subfigure]{font=scriptsize,labelformat=empty,justification=centering}
  \includegraphics[scale=0.48]{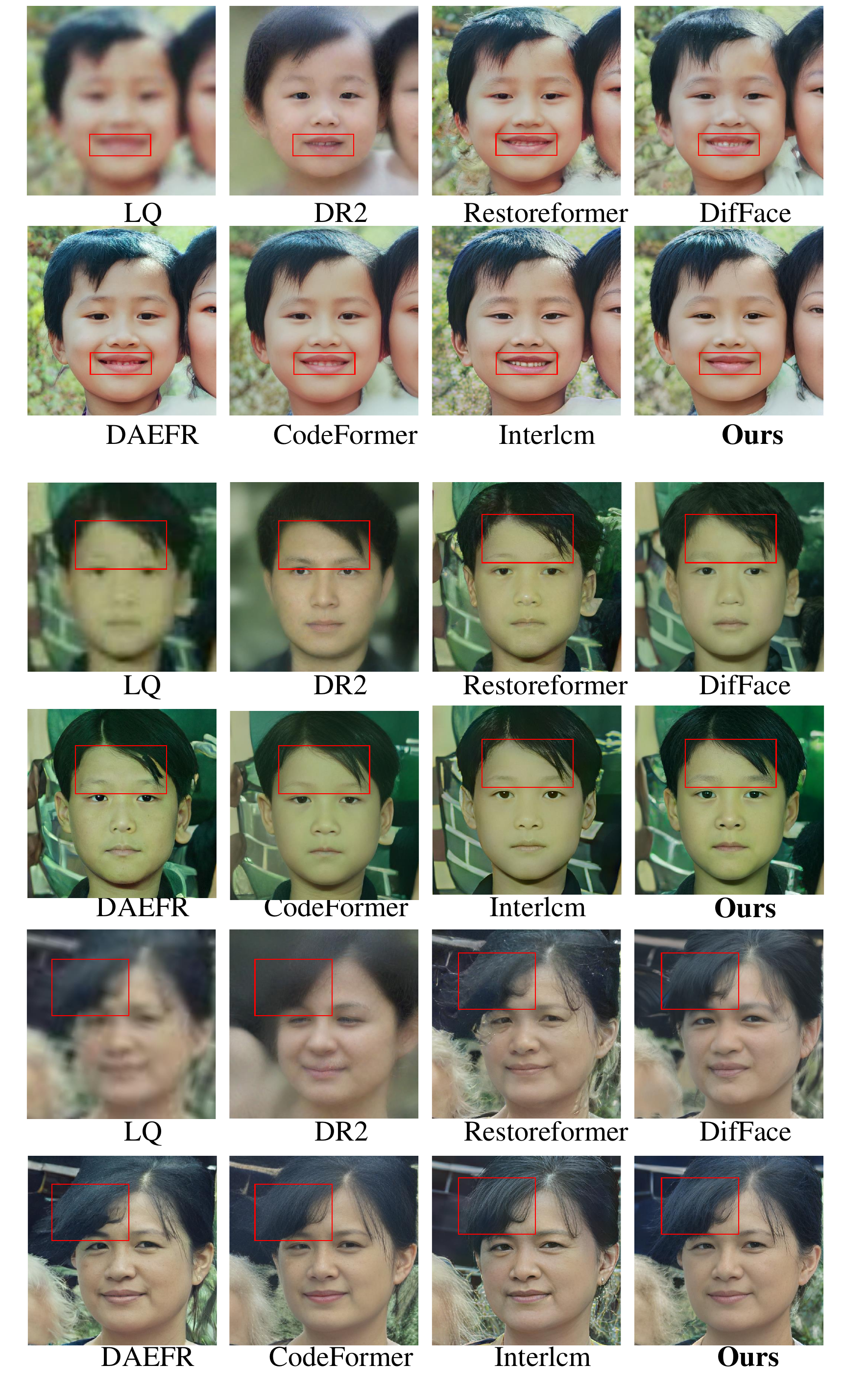}
  \caption{The qualitative comparisons on the WebPhoto-Test. Our results are sharper and have more facial details (\eg, hairs and eyebrows).  Please zoom in for the best view.}
  \label{Fig.vis_web}
\end{figure*}

\begin{figure*}[t]
  \setlength{\abovecaptionskip}{-0.4cm}
    \centering
  \captionsetup[subfigure]
{font=scriptsize,labelformat=empty,justification=centering}
  \includegraphics[scale=0.48]{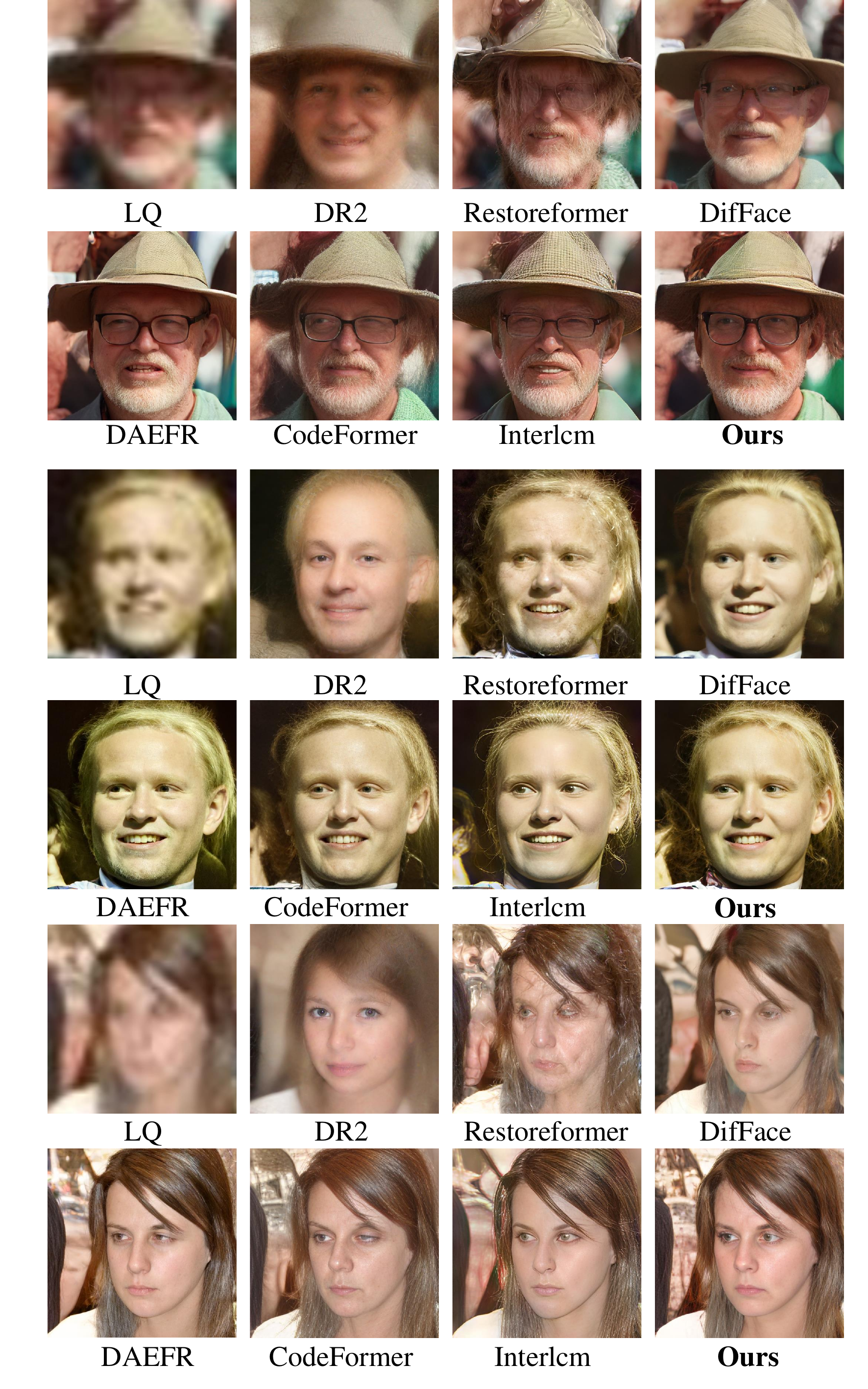}

  \caption{The qualitative comparisons on the WIDER-Test. Our results have more facial details and less artifact.  Please zoom in for the best view.}
  \label{Fig.vis_wider}
\end{figure*}

In this section provide more visual comparisons with state-of-the-art methods, 
including Interlcm\cite{li2025interlcm}, CodeFormer\cite{codeformer}, DifFace\cite{yue2022difface}, Restoreformer \cite{wang2022restoreformer}, DR2\cite{wang2023dr2}, DAEFR\cite{tsai2023dual}.
\Cref{Fig.vis_celeba} shows the qualitative comparisons on the CelebA-Test.
\Cref{Fig.vis_lfw} shows the qualitative comparisons on the LFW-Test.
\Cref{Fig.vis_web} shows the qualitative comparisons on the WebPhoto-Test.
\Cref{Fig.vis_wider} shows the qualitative comparisons on the WIDER-Test.
Generally, our results have higher sharpness, richer textures (e.g., eyebrows, lips), and are robust to varying degrees of degradation.
\label{sec:6.5}

